\PassOptionsToPackage{table}{xcolor}
\documentclass{article}

\usepackage{iclr2027/iclr2027_conference,times}
\usepackage{amsmath,amsfonts,bm}

\def\eqref#1{equation~\ref{#1}}
\def\1{\bm{1}}

\DeclareMathAlphabet{\mathsfit}{\encodingdefault}{\sfdefault}{m}{sl}
\SetMathAlphabet{\mathsfit}{bold}{\encodingdefault}{\sfdefault}{bx}{n}

\usepackage{hyperref}
\usepackage{url}
\usepackage{booktabs}
\usepackage{graphicx}
\usepackage{multirow}
\usepackage{flafter}
\usepackage{placeins}
\usepackage{wrapfig}

\newcommand{\methodname}{\textsc{MAGIC}}

\title{\methodname{}: Mixed-Granularity Agent Graphs via Incremental
Construction with Dense-Reward Reinforcement Learning}
\author{\textbf{Kairui Yang \quad Ziheng Yi \quad Xunkai Li \quad Minghao An} \\ \textbf{Zhanke Liu \quad Zekai Chen \quad Rong-Hua Li}}

\iclrfinalcopy

\begin{document}

\maketitle
\lhead{Preprint}
\hypersetup{hidelinks,pdfauthor={Kairui Yang, Ziheng Yi, Xunkai Li, Minghao An, Zhanke Liu, Zekai Chen, Rong-Hua Li},pdftitle={MAGIC: Mixed-Granularity Agent Graphs via Incremental Construction with Dense-Reward Reinforcement Learning}}

\begin{abstract}
Collaboration topology shapes both the performance and execution cost of
LLM-based multi-agent systems. Because tasks differ in complexity and required
capabilities, recent approaches generate task-specific collaboration graphs that
specify agent participation and information flow. However, representative topology
generators use either individual agents or predefined groups throughout an
organization, overlooking differing collaboration needs across subtasks. Our key
insight is to select granularity locally for each functional role, combining
fine-grained control with reusable collaboration patterns within one organization.
Learning such organizations requires exploring a combinatorial construction space
with limited intermediate
feedback from final-answer rewards. Therefore, we propose \methodname{}, a dense-reward
reinforcement learning framework for mixed-granularity graph generation.
Specifically, \methodname{} constructs a mixed-granularity agent graph by sequentially selecting
a functional role, instantiating it as a single agent or reusable group, and
connecting it to existing units. We directly optimize the construction policy
using returns from trajectories sampled under the current policy and use
potential-based reward shaping to provide intermediate feedback from probe-based
utility and structural signals while preserving the cumulative task reward.
\methodname{} outperforms state-of-the-art baselines across eight benchmarks
and demonstrates strong inference efficiency in our efficiency study.
\end{abstract}

\section{Introduction}

\begin{wrapfigure}{r}{0.48\textwidth}
    \centering
    \includegraphics[width=\linewidth]{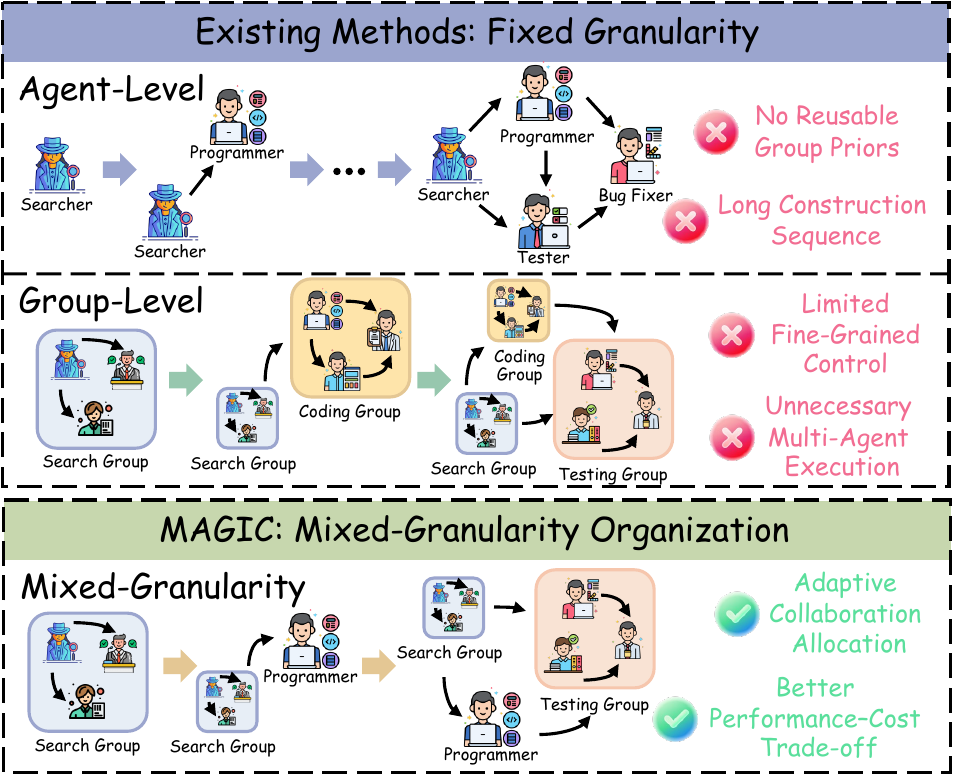}
    \caption{Fixed-granularity organizations versus \methodname{}'s
    mixed-granularity construction for an API-update task.}
    \label{fig:granularity_gap}
\end{wrapfigure}
LLM-based multi-agent systems (MAS) coordinate specialized language-model
agents through structured interaction
\citep{wu2024autogen,du2023improving,liu2023dylan}. Their effectiveness and
cost depend on collaboration topology, which determines agent participation
and information flow
\citep{zhuge2024gptswarm,li2024sparse,zhang2025gdesigner}. Because tasks differ in
complexity and required capabilities, recent work generates these topologies
conditioned on each task \citep{zhang2025gdesigner,li2025argdesigner,chen2026goagent}.
The closest approaches follow two routes: agent-level methods predict edges
over specified agents or generate atomic-agent roles and connections
\citep{zhang2025gdesigner,li2025argdesigner}, whereas group-level methods
connect predefined collaborative groups as reusable units
\citep{chen2026goagent}. Despite their different scales, both fix node
granularity before construction, producing only atomic-agent or only group
topologies.

As illustrated in Figure~\ref{fig:granularity_gap}, this shared choice creates
a \emph{fixed-granularity mismatch}: subtasks within
one query can demand different amounts of collaboration, yet every constructed
unit must use the same organizational scale. Consider an API-update task.
Investigating upstream changes and testing compatibility may merit a
\texttt{SearchGroup} and a \texttt{TestingGroup}, while implementing a one- or
two-line patch may need only a \texttt{ProgrammerAgent}. The resulting gap
concerns reusable collaboration priors, construction length, and execution
cost: an all-atomic graph must assemble collaboration node by node, without
reusable group priors \citep{chen2026goagent} and with longer construction sequences, whereas an
all-group graph treats each group as an indivisible construction unit, limiting
fine-grained structural control and thereby imposing unnecessary multi-agent
execution on subtasks that a single agent can handle. The controlled assignments in
Section~\ref{sec:empirical_investigation} (D1) show task- and role-dependent
performance--cost preferences, motivating local granularity selection in a
mixed space whose all-atomic and all-group organizations remain special cases.

Learning in this mixed space requires an effective optimization route and
informative feedback for intermediate construction decisions. Search-then-SFT
trains generators to imitate construction trajectories selected through search
and verification \citep{li2025argdesigner,chen2026goagent}. Mixed granularity
adds configurations to this search: even with a fixed role--edge skeleton,
$m$ dual-granularity roles admit $2^m$ realizations. D2 bounds the fraction
verifiable under a given execution budget. We study an alternative learning
route that uses returns from current-policy trajectories directly for updates,
without first selecting demonstration graphs. Section~\ref{sec:matched_optimization}
compares these routes under matched API cost budgets. Separately, trajectories
sampled for the same query receive identical terminal-only returns when their
final scores coincide. D3 shows that potential-based shaping can distinguish
some such trajectories at intermediate construction steps by incorporating
information about their partial graphs.
This motivates complementing return-based construction learning with
intermediate graph feedback.

We therefore propose \methodname{}, a dense-reward reinforcement learning
framework for mixed-granularity graph generation through incremental
construction of an agent graph from an empty graph. At each
construction step, a task-conditioned policy selects a functional role, its
atomic-or-group realization, and predecessors, yielding an executable
organization. The mixed-granularity organization space permits local
atomic-or-group choices, and training
uses proposals sampled from the current policy without a pre-collected corpus
of successful trajectories. We use potential-based reward shaping to provide
construction-level feedback from probe utility, structural complexity, and
role repetition while preserving the terminal task objective.

\textbf{Our contributions}. (1) \textbf{Mixed-granularity perspective.} We formulate task-conditioned
MAS topology generation in a mixed-granularity organization space that
supports local selection between single agents and reusable groups, with
all-atomic and all-group organizations as special cases. (2) \textbf{Dense-reward incremental construction.} We propose
\methodname{}, which directly optimizes role, granularity, and dependency
decisions using returns from current-policy trajectories with
potential-based construction feedback. (3) \textbf{Observed evidence.} \methodname{} outperforms
state-of-the-art baselines across eight benchmarks and demonstrates strong
inference efficiency in our efficiency study.

\section{Problem Formulation}
\label{sec:problem_formulation}

Given a task query $\mathcal Q$, we construct a task-specific multi-agent
organization from a fixed domain library $\mathcal R$. Its role profiles and
admissible realizations are built once offline and held fixed across queries;
construction changes only the organization selected for $\mathcal Q$.

\paragraph{Mixed-granularity organization space.}
Each library role $r\in\mathcal R$ admits one or both realizations in
$\mathcal Z(r)\subseteq\{\mathsf{AtomicAgent},\mathsf{Group}\}$. A
task-specific mixed-granularity agent graph is
\begin{equation}
    \mathcal G=(\mathcal U,\mathcal E),\qquad
    \mathcal U=\{u_i=(i,r_i,z_i)\}_{i=1}^{N},\qquad
    z_i\in\mathcal Z(r_i),
    \label{eq:mixed_graph}
\end{equation}
where $\mathcal U$ contains top-level units and $\mathcal E\subseteq
\mathcal U\times\mathcal U$ their directed dependencies. An atomic unit
executes its role directly; a group is a fixed internal agent graph with the
same external interface. The mixed space $\mathbb G_{\mathrm{mix}}$ contains
the restricted all-atomic and all-group spaces $\mathbb G_{\mathrm{atomic}}$
and $\mathbb G_{\mathrm{group}}$.

\paragraph{Task objective and constraints.}
Let $\widehat y=\operatorname{Execute}(\mathcal Q,\mathcal G)$ denote the
answer obtained by executing organization $\mathcal G$ on query $\mathcal Q$,
with execution specified in Section~\ref{sec:method_graph}. Let
$S(\widehat y,y^*)$ be its score against the reference answer or tests $y^*$.
We seek a construction policy inducing
$p_\theta(\mathcal G\mid\mathcal Q,\mathcal R)$ that maximizes expected task
score:
\begin{equation}
    \max_\theta
    \mathbb E_{(\mathcal Q,y^*)}
    \mathbb E_{\mathcal G\sim p_\theta(\cdot\mid\mathcal Q,\mathcal R)}
    \left[
      S(\operatorname{Execute}(\mathcal Q,\mathcal G),y^*)
    \right],
    \label{eq:problem_objective}
\end{equation}
subject to $\mathcal G\in\mathbb G_{\mathrm{legal}}$ and
$N\leq N_{\max}$. Here $\mathbb G_{\mathrm{legal}}\subseteq
\mathbb G_{\mathrm{mix}}$ requires admissible role realizations, compatible
interfaces, and acyclic expanded graphs; $N_{\max}$ bounds the number of
top-level units. Execution cost is evaluated alongside task performance.
Gold answers or tests are used only after execution for scoring.
The space raises questions about local granularity, configuration verification, and
construction credit, examined next.

\section{Empirical Investigation}
\label{sec:empirical_investigation}

We examine local granularity preferences (D1), configuration verification
costs (D2), and intermediate construction feedback (D3).

\begin{figure*}[!t]
    \centering
    \includegraphics[width=\textwidth]{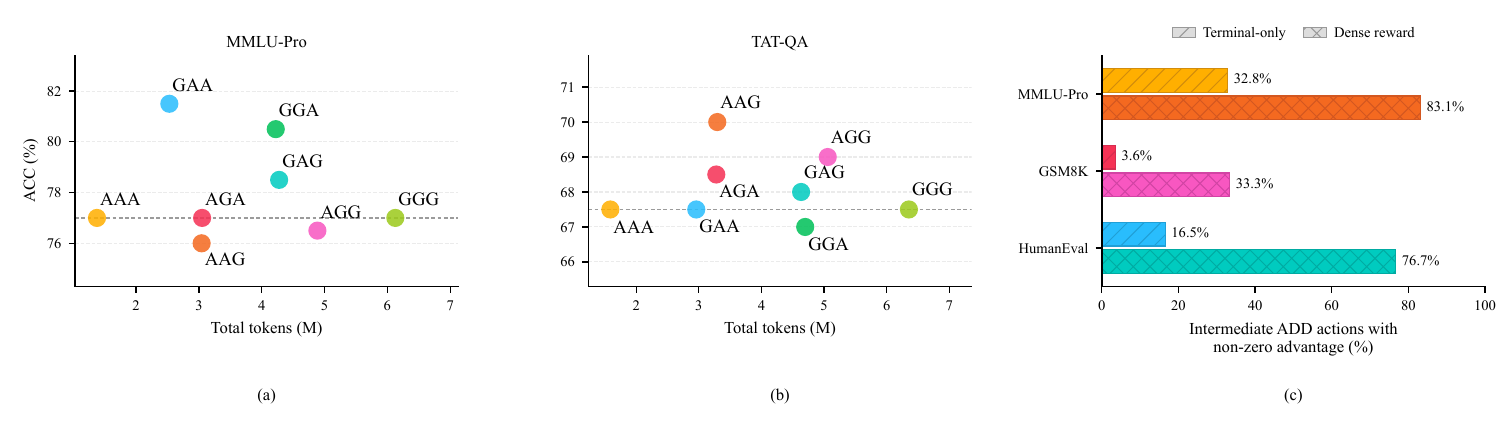}
    \caption{\textbf{Granularity preferences and construction feedback.}
    (a,b) D1: accuracy versus total input/output tokens on MMLU-Pro and TAT-QA.
    Letters denote atomic (A) or group (G) realizations of the decomposer,
    solver, and verifier. TAT-QA uses the diagnostic's strict-answer protocol.
    (c) D3: non-zero advantage rates for intermediate ADD actions,
    excluding initial decisions and STOP.}
    \label{fig:empirical_investigation}
\end{figure*}

\paragraph{D1: Granularity preferences are task and role dependent.}
With a fixed library and execution settings, we enumerate all eight
atomic/group assignments of a serial decomposer--solver--verifier skeleton
on 200 queries each from MMLU-Pro test and TAT-QA dev.
Figure~\ref{fig:empirical_investigation}(a,b) shows that MMLU-Pro favors
\texttt{GAA}, whereas TAT-QA favors \texttt{AAG}. These mixed assignments
outperform both fixed-granularity endpoints by 4.5 and 2.5 accuracy points,
respectively, while consuming fewer tokens than \texttt{GGG}.
The preferred grouped role thus differs by task, supporting local granularity
selection. Appendix~\ref{app:diagnostic_d1} gives the protocol and detailed
comparisons.

\paragraph{D2: Verified configuration coverage is analytically bounded.}
For a fixed role--edge skeleton with $m$ dual-granularity roles and a budget
of $B$ complete graph executions, the configuration count and verifiable
fraction satisfy
\begin{equation}
    |\mathcal Z_{\mathrm{mix}}|=2^m,\qquad
    \operatorname{Coverage}(B)\leq\min\!\left(1,\frac{B}{2^m}\right),
    \label{eq:supervision_coverage_bound}
\end{equation}
since verifying each assignment requires at least one execution. Role and
edge choices further enlarge the construction space.

\paragraph{D3: Relative learning signals at intermediate construction steps.}
On the same 720 trajectories from MMLU-Pro, GSM8K, and HumanEval, we compare
terminal-only feedback with the dense reward defined in
Section~\ref{sec:method_optimization}.
Figure~\ref{fig:empirical_investigation}(c) shows higher non-zero advantage
rates for intermediate ADD actions on all three datasets, including an
increase from 32.81\% to 83.13\% on MMLU-Pro. Dense returns also distinguish
some equal-terminal-score trajectories at intermediate positions, supporting
partial-graph information as a source of relative learning signals.
Appendix~\ref{app:diagnostic_d3} reports the protocol and full statistics;
Section~\ref{sec:component_ablations} evaluates the learning benefits of this
feedback.

These observations motivate local granularity choices and intermediate
feedback. Section~\ref{sec:matched_optimization} compares direct policy
optimization with Search-then-SFT under matched API cost budgets.

\FloatBarrier

\section{Methodology}
\label{sec:methodology}

\begin{figure}[!t]
    \centering
    \includegraphics[width=\linewidth]{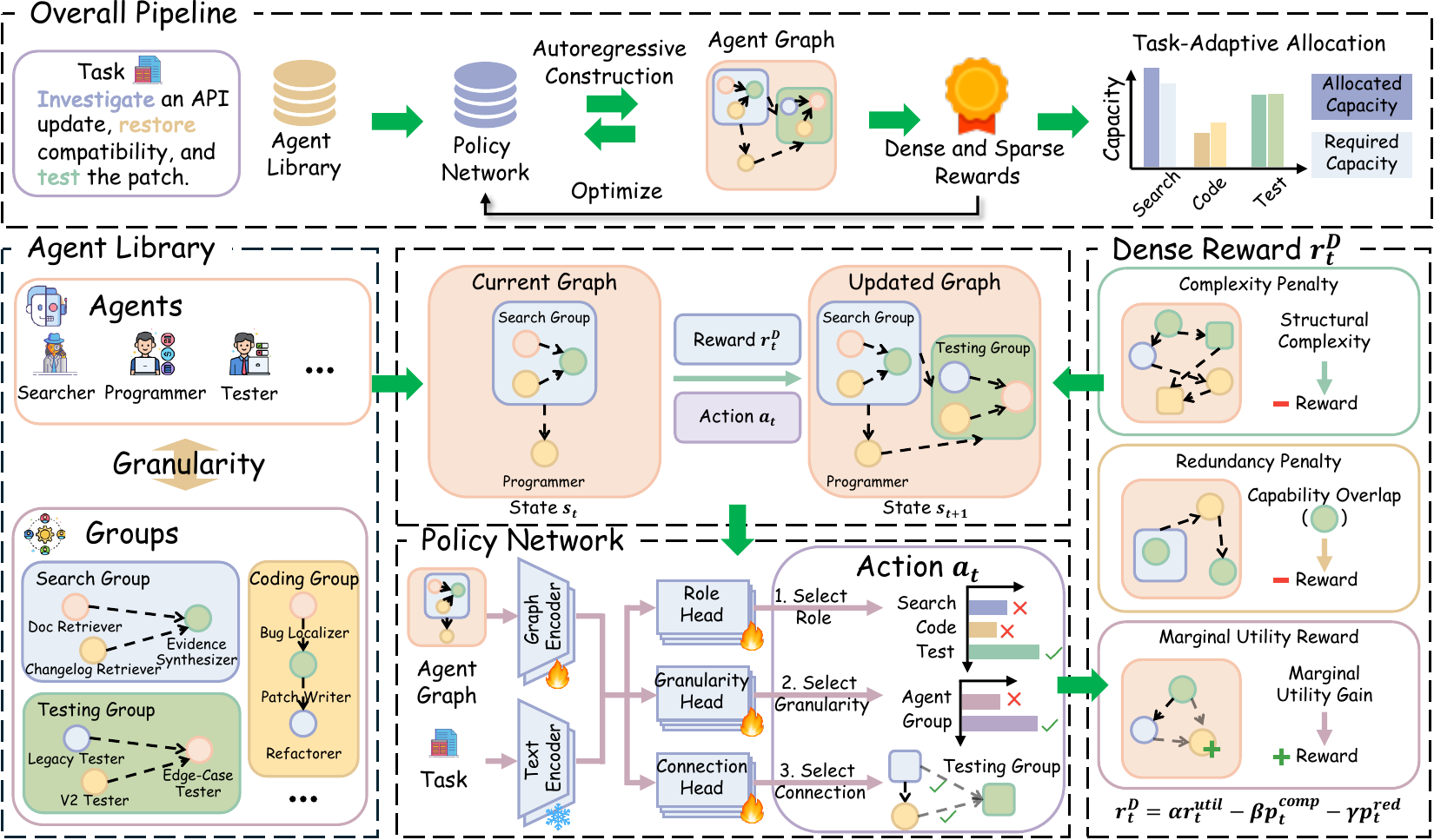}
    \caption{Overview of the proposed \methodname{} framework.}
    \label{fig:method_framework}
\end{figure}

As illustrated in Figure~\ref{fig:method_framework}, \methodname{} learns to
construct an executable mixed-granularity agent graph from a task query and a fixed
role library. A task-conditioned policy incrementally selects roles,
atomic-or-group realizations, and dependencies. Frozen LLM executors then
execute the resulting organization. Training uses on-policy trajectories
and potential-based reward shaping to provide construction-level feedback
while preserving the terminal task objective.

\subsection{Mixed-Granularity Realization and Execution}
\label{sec:method_graph}

We instantiate the units defined in Section~\ref{sec:problem_formulation}
with frozen LLM executors. An atomic unit uses one role executor; a group
uses its internal agents and an aggregator that emits a single role-level
result. Both realizations expose this result through the same input--output
contract, allowing connected units to exchange role-level outputs regardless
of granularity. Internal group membership and topology remain fixed during
construction.

To execute a nonempty partial or final graph, we expand each group into its
internal DAG and run the resulting atomic-agent graph in dependency order.
A fixed outer summarizer produces the final answer from the public task,
output requirements, and results from all sinks of the expanded graph.
For the empty graph $\mathcal G_0$, including immediate \textsc{stop},
the base LLM instead receives a minimal direct-answer prompt containing the
question, any answer options, and the answer requirement.
Appendix~\ref{app:policy_architecture} details the execution semantics, and
Appendix~\ref{app:role_protocols} specifies the role library and interfaces.

\subsection{Task-Conditioned Incremental Construction}
\label{sec:method_construction}

The policy observes the public query and partial agent graph. A frozen text encoder
embeds the query and role realizations; node features combine realization
semantics with granularity, construction position, and degree embeddings.
An \emph{Edge-aware GRU} encodes units in construction order, combining each
unit's features with the mean representation of its actual predecessors and
the query embedding. Its recurrent state captures construction history,
whereas predecessor aggregation captures graph dependencies.
Task-conditioned attention pools the node states, and a projection combines
the pooled representation, query, and construction progress into decision
context $\mathbf c_t$. A learned empty-graph vector initializes this process.
Appendix~\ref{app:construction_policy} gives the encoder and action-head
formulations.

Three conditioned heads parameterize a non-\textsc{stop} action
$a_t=(r_t,z_t,\mathbf b_t)$, where $b_{jt}$ indicates an edge from existing
unit $j$ to the new unit:
\begin{equation}
    \pi_\theta(a_t\mid s_t)
    =\pi_\theta(r_t\mid s_t)
     \pi_\theta(z_t\mid r_t,s_t)
     \prod_{j\in\mathcal V_t}\pi_\theta(b_{jt}\mid r_t,z_t,s_t).
    \label{eq:graph_generation_heads}
\end{equation}
Here $\mathcal V_t$ contains legal predecessors. Roles and \textsc{stop}
share one masked categorical distribution; a second categorical distribution
selects the chosen role's realization. Independent Bernoulli decisions then
select an admissible subset of predecessors. A \textsc{stop} action uses
only its role-level probability and skips the remaining heads. Legality masks
enforce the library, direction, and structural constraints.

Each addition updates the graph encoding before the next decision.
Construction ends with an explicit \textsc{stop} action; at the maximum unit
horizon, all other roles are masked, making \textsc{stop} the only legal choice.
At inference,
the policy samples a graph with dropout disabled and then executes it;
probe scores, gold answers, and rewards are training-only information.
The text encoder and execution LLMs remain frozen throughout training; only
the graph encoder, feature projections, pooling, and action heads are updated.

\subsection{Potential-Based Dense-Reward Optimization}
\label{sec:method_optimization}

\paragraph{Task objective and graph potential.}
Let $S(\mathcal G,q)\in[0,1]$ denote the score obtained by executing graph
$\mathcal G$ on query $q$. Training maximizes expected terminal task score,
using accuracy, official F1, or code-test success as appropriate.
For each query's trajectory group, we select a fixed probe set $P$ from other
training queries and share it across all trajectories and construction steps.
Executing partial graphs on these probes provides a common, multi-query
reference for their utility, complementing the terminal score on the
conditioning query. Keeping $P$ fixed lets us compare successive organizations
on the same questions. Adding a unit changes the graph from $\mathcal G_t$ to
$\mathcal G_{t+1}$, yielding the paired utility gain:
\begin{equation}
    \Delta U_t=
    \frac{\sum_{p\in P}[S(\mathcal G_{t+1},p)-S(\mathcal G_t,p)]}
    {|P|+\kappa}.
    \label{eq:paired_utility_gain}
\end{equation}
Here $\kappa\geq0$ smooths the gain. To represent structural overhead, we define
$C(\mathcal G)=w_n[N_{\mathrm{exp}}(\mathcal G)-b_n]_+
+w_e[E_{\mathrm{exp}}(\mathcal G)-b_e]_+$, where $[x]_+=\max(0,x)$,
$N_{\mathrm{exp}}$ and $E_{\mathrm{exp}}$ count nodes and edges after
expanding groups, and $b_n,b_e$ are free allowances.
Role repetition is $D(\mathcal G)=\sum_r[m_r(\mathcal G)-1]_+$, where
$m_r$ counts outer units assigned role $r$, regardless of their granularity.
Thus, each addition contributes
$d_t=\alpha\Delta U_t-\beta\Delta C_t-\gamma_R\Delta D_t$,
with $\Delta C_t=C(\mathcal G_{t+1})-C(\mathcal G_t)$ and analogously for
$\Delta D_t$. The weights balance utility gains against added structural
complexity and repeated roles in the shaping signal.
Starting from $\Phi_0=0$, we accumulate $\Phi_{t+1}=\Phi_t+d_t$;
fixed probes and consistent graph scores define the graph potential
$\Phi_t=\Phi_P(\mathcal G_t)$ relative to the initial graph.
Evaluation caching and full reward definitions are detailed in
Appendix~\ref{app:reward_specification}; Appendix~\ref{app:configuration}
records the documented configuration.

\paragraph{Shaping and terminal settlement.}
With discount $\eta$ shared by shaping and return computation, the reward is
\begin{equation}
    r'_t=r_t^{\mathrm{task}}+\eta\Phi_{t+1}-\Phi_t.
    \label{eq:dense_transition_reward}
\end{equation}
The base reward is zero during construction and the task score at termination.
We set initial and terminal potentials to zero. Thus, explicit \textsc{stop}
receives $S(\mathcal G,q)-\Phi_P(\mathcal G)$, including when forced by the
unit horizon. The implementation accumulates
paired utility and structural increments to compute the intermediate
potential. With the current $\eta=1$, shaping redistributes reward across
construction steps while the trajectory total remains its terminal score.

\paragraph{Policy optimization.}
For each query, we sample $K$ trajectories from the current policy and compute
discounted return-to-go $G_{k,t}$. Returns are normalized among trajectories
that contain an action at the same position: $A_{k,t}=(G_{k,t}-\mu_t)/
(\sigma_t+\epsilon)$. We minimize the action-averaged loss
\begin{equation}
    \mathcal L
    =-\overline{\operatorname{sg}(A)\log\pi_\theta(a\mid s)}
     +\beta_{\mathrm{KL}}\overline{D_{\mathrm{KL}}(\pi_\theta\Vert\pi_{\mathrm{ref}})}
     -\beta_H\overline{H(\pi_\theta)}.
    \label{eq:policy_loss}
\end{equation}
The bar averages over all recorded action positions, including the final
\textsc{stop} even when forced by the horizon. This probability-one action
has zero log-probability but remains in position-aligned normalization and
the averaging denominator. The operator $\operatorname{sg}$ stops gradients
through advantages.
The reference policy is a frozen copy from the start of RL. KL and entropy
are computed over the full hierarchical action distribution at each visited
state. This trains the construction policy directly from its own trajectories,
without a pre-collected successful-trajectory corpus or a value network.
Appendix~\ref{app:grpo_objective} specifies normalization and regularization.

Training and inference incur different execution costs. During training,
multiple trajectories per query use partial-graph evaluations on shared
probes and final-graph scoring on the conditioning query, followed by return
computation and policy updates. At inference, the policy constructs a graph
from the public query and executes only the completed organization. Probe
evaluation and policy updates are confined to training
(Appendix~\ref{app:training_inference_algorithms}).

\subsection{Policy Invariance of Reward Shaping}
\label{sec:shaping_theory}

Following potential-based reward shaping~\citep{ng1999policy}, we establish
the relation between the shaped and terminal task objectives.
\textbf{Proposition 1 (Return preservation).}
For a finite construction episode with $T$ reward transitions, use the same
discount $\eta$ in shaping and returns, set $\Phi_0=\Phi_T=0$, and fully
settle the potential under both sampled and forced termination. Then
\begin{equation}
    \sum_{t=0}^{T-1}\eta^t r'_t
    =\sum_{t=0}^{T-1}\eta^t r_t^{\mathrm{task}}.
    \label{eq:shaping_invariance}
\end{equation}
Consequently, shaping preserves the expected task return and its optimal
policy set. In our undiscounted setting, both returns equal the final task
score. The proof telescopes the potential differences and is given in
Appendix~\ref{app:shaping_proof}.

This property separates the task objective from construction-level feedback:
probe utility and structural information redistribute learning signals without
adding a lasting cost term to the terminal objective. For $\eta=1$, the
shaped return-to-go is $G_{k,t}=S(\mathcal G_k,q)-\Phi_{k,t}$.
The pre-action potential serves as a state-dependent reference for the
organization already built. Position-aligned normalization therefore compares
terminal outcomes relative to these partial-graph potentials, yielding the
intermediate signal differences examined in D3 even when final scores coincide.
The proposition concerns environment returns; policy updates use the
normalized, KL- and entropy-regularized loss in Equation~\ref{eq:policy_loss}.

\section{Experiments}
\label{sec:experiments}

We organize the evaluation around four questions. \textbf{Q1:Effectiveness.} Does
\methodname{} consistently outperform strong baselines across four task
families and eight benchmarks (Sec.~\ref{sec:overall_performance})?
\textbf{Q2:Interpretability.} Do mixed granularity, direct graph-generation optimization, and
dense construction credit each contribute to the learned performance--cost
trade-off (Sec.~\ref{sec:component_ablations})? \textbf{Q3: Optimization Strategies.} Under matched
Qwen Flash API cost budgets, how does the dense-return
policy-gradient training route compare with Search-then-SFT
(Sec.~\ref{sec:matched_optimization})? \textbf{Q4:Efficiency.} Does \methodname{} achieve
a favorable inference performance--token trade-off relative to alternative
agent organizations (Sec.~\ref{sec:performance_cost_tradeoff})?

\subsection{Experimental Setup}
\label{sec:experimental_setup}

\paragraph{Datasets.}
We evaluate \methodname{} on eight benchmarks spanning four task families:
(1) knowledge and commonsense reasoning: MMLU-Pro~\citep{wang2024mmlupro} and
StrategyQA~\citep{geva2021strategyqa}; (2) mathematical reasoning:
AQuA~\citep{ling2017program} and GSM8K~\citep{cobbe2021training}; (3) code
generation: HumanEval~\citep{chen2021evaluating} and
LiveCodeBench-v6 (LCB-v6)~\citep{jain2025livecodebench}; and (4) table and financial
reasoning: TAT-QA~\citep{zhu2021tatqa} and TabFact~\citep{chen2020tabfact}.
Data and scoring protocols are detailed in Appendix~\ref{app:datasets_scoring}.

\paragraph{Baselines.}
We compare against 17 baselines in three categories:
(1) direct/single-agent methods: DeepSeek V4 Flash~\citep{deepseek2026v4flash}
and Qwen Flash~\citep{alibaba2026qwenflash};
(2) established MAS: AutoGen~\citep{wu2024autogen} and
LLM-Debate~\citep{du2023improving}; and
(3) adaptive/learned organizations: DyLAN~\citep{liu2023dylan},
GPTSwarm~\citep{zhuge2024gptswarm}, G-Designer~\citep{zhang2025gdesigner},
Sparse-Comm~\citep{li2024sparse}, GraphSearch~\citep{wu2024graphsearch},
Graph-R1~\citep{luo2025graphr1}, R-GFM~\citep{liu2026rgfm},
BIGMAS~\citep{hao2026bigmas}, MasRouter~\citep{yue2025masrouter},
GoAgent~\citep{chen2026goagent}, VeriMAP~\citep{xu2026verimap},
ARG-Designer~\citep{li2025argdesigner}, and EIB-Learner~\citep{shen2025eib}.
Baseline provenance and implementation settings are provided in
Appendices~\ref{app:baseline_protocols} and~\ref{app:configuration}.

\subsection{Overall Performance}
\label{sec:overall_performance}

For Q1, \methodname{} uses DeepSeek V4 Flash as its executor. We train a
separate construction policy for each benchmark on 100 training examples;
HumanEval uses the same 100 examples as LiveCodeBench. Each benchmark uses
one training run, and we evaluate its final checkpoint. Dataset-specific
training and evaluation partitions are detailed in
Appendix~\ref{app:q1_q4_protocol}.

\begin{table*}[!ht]
  \centering
  \small
  \renewcommand{\arraystretch}{1.20}
  \caption{Overall performance across eight benchmarks (\%). We report
  official F1 for TAT-QA, pass@1 for code-generation tasks, and accuracy
  for all remaining tasks.
  Best scores in each column are shown in \textbf{bold}.}
  \label{tab:main_results}
  \setlength{\tabcolsep}{2pt}
  \resizebox{\textwidth}{!}{%
  \begin{tabular}{c|*{3}{*{2}{>{\centering\arraybackslash}m{5.2em}}|}*{2}{>{\centering\arraybackslash}m{5.2em}}}
    \specialrule{1.5pt}{1.5pt}{1.5pt}
    {}
    & \multicolumn{2}{c|}{\begin{tabular}{@{}c@{}}\textbf{Knowledge \&} \\ \textbf{Commonsense} \\ \textbf{(Acc. \%)}\end{tabular}}
    & \multicolumn{2}{c|}{\begin{tabular}{@{}c@{}}\textbf{Mathematical} \\ \textbf{Reasoning} \\ \textbf{(Acc. \%)}\end{tabular}}
    & \multicolumn{2}{c|}{\begin{tabular}{@{}c@{}}\textbf{Code} \\ \textbf{Generation} \\ \textbf{(pass@1 \%)}\end{tabular}}
    & \multicolumn{2}{c}{\begin{tabular}{@{}c@{}}\textbf{Table \& Financial} \\ \textbf{Reasoning} \\ \textbf{(F1 / Acc. \%)}\end{tabular}} \\
    \cmidrule(lr){2-3}\cmidrule(lr){4-5}\cmidrule(lr){6-7}\cmidrule(lr){8-9}
    \textbf{Method} & \textbf{MMLU-Pro} & \textbf{StrategyQA} & \textbf{AQuA} & \textbf{GSM8K} & \textbf{HumanEval} & \textbf{LCB-v6} & \textbf{TAT-QA} & \textbf{TabFact} \\
    \midrule

      DeepSeek V4 Flash
      & 72.80 & 85.40
      & 89.76 & {93.71}
      & 84.76 & 20.57
      & 70.48 & 91.60 \\
      Qwen Flash
      & 60.60 & 78.00
      & 68.90 & 59.44
      & {90.24}
      & {26.86}
      & 69.51 & 80.00 \\

      AutoGen
      & 73.00 & 85.20
      & 88.98 & 92.87
      & 81.71 & 18.86
      & 70.29 & 92.00 \\
      LLM-Debate
      & 72.20 & 84.40
      & 89.37 & 92.87
      & 83.54 & 21.71
      & 70.66 & {92.80} \\

      DyLAN
      & {73.60}
      & 83.00 & 89.76
      & 93.33 & 85.98
      & 22.86 & 70.48
      & {93.80} \\

      GPTSwarm
      & 71.20 & 84.60
      & {91.34}
      & 93.56 & 84.76
      & 19.43 & 70.41
      & 92.00 \\

      G-Designer
      & 70.20
      & {87.00}
      & 88.98 & 93.63
      & 85.37 & {23.43}
      & 70.23 & 90.80 \\

      Sparse-Comm
      & 70.60 & 85.00
      & 89.76
      & {93.86}
      & 84.76 & 21.14
      & 70.41 & 92.20 \\

      GraphSearch
      & 71.60 & {85.80}
      & 89.37 & 93.56
      & 81.10 & 22.29
      & {71.26}
      & 91.80 \\

      Graph-R1
      & 71.60 & 84.20
      & {90.55} & 92.87
      & 86.59 & 20.00
      & 69.75 & 92.00 \\

      R-GFM
      & 68.60 & 84.20
      & 89.76 & 93.18
      & 89.02 & 19.43
      & {71.20} & 90.60 \\

      BIGMAS
      & 69.80 & 84.40
      & {90.55} & 93.18
      & 84.76 & 21.71
      & 70.41 & 92.40 \\

      MasRouter
      & 71.20 & 83.40
      & 88.98 & {93.71}
      & 76.83 & 20.57
      & 69.99 & 92.60 \\

      GoAgent
      & 70.80 & 84.00
      & 89.76 & 92.87
      & 79.88 & 20.57
      & 71.14 & 91.40 \\

      VeriMAP
      & 72.40 & 82.00
      & 89.76 & 93.63
      & 78.66 & 21.14
      & 70.23 & 91.80 \\

      ARG-Designer
      & {73.40} & 83.20
      & {90.55} & 93.63
      & {89.63} & 19.43
      & 71.02 & 92.00 \\
      EIB-Learner
      & 72.00 & 83.80
      & 89.76 & 93.56
      & 83.54 & 21.71
      & 69.99 & 91.60 \\
    \cmidrule(lr){1-9}
    \methodname{} (Ours)
      & {\textbf{83.20}}
      & {\textbf{87.20}}
      & {\textbf{91.73}}
      & {\textbf{94.92}}
      & {\textbf{90.85}}
      & {\textbf{36.57}}
      & {\textbf{72.22}}
      & {\textbf{94.00}} \\
    \specialrule{1.3pt}{2.0pt}{1.0pt}
  \end{tabular}%
  }
\end{table*}

To answer \textbf{Q1}, Table~\ref{tab:main_results} compares \methodname{}
with 17 baselines across four task families. \methodname{} ranks first on
all eight benchmarks, outperforming direct inference, established MAS, and
adaptive organizations. The largest gains over the strongest baselines occur
on MMLU-Pro and LCB-v6: accuracy increases from 73.60\% to 83.20\%, while
pass@1 rises from 26.86\% to 36.57\%. Improvements also extend to mathematical
and structured-data reasoning, where several baselines already obtain
closely clustered scores. Thus, the overall advantage spans both reasoning
accuracy and executable-code correctness.

The leading baseline varies across tasks, highlighting the importance of
task-dependent organization. In particular, Qwen Flash outperforms every MAS
baseline on both code-generation benchmarks, yet \methodname{} improves
further on both. On MMLU-Pro and TAT-QA, it instead surpasses the learned
organizations DyLAN and GraphSearch, respectively. This pattern is consistent
with the motivation for adapting collaboration to each task rather than
using a uniform execution scale. \methodname{} supports such adaptation through
local atomic-or-group choices and feedback-driven construction learning.
Together, these results establish the framework's effectiveness across
heterogeneous tasks; the following ablations examine the contributions of
its organization space and training design.

\subsection{Component Ablations}
\label{sec:component_ablations}

To answer \textbf{Q2}, we ablate granularity selection, policy optimization,
and dense feedback under shared Qwen Flash settings on MMLU-Pro and TAT-QA.
All-Atomic and All-Group restrict training and inference to one granularity;
Initialization only evaluates the untrained policy; Final-reward only keeps
the mixed space and RL updates but removes potential-based shaping, using
only terminal task scores.

\begin{table*}[!ht]
  \centering
  \small
  \caption{Component ablations on MMLU-Pro and TAT-QA.
  Best scores and lowest token counts are shown in
  \textbf{bold}; second-best values are underlined.
  Arrows indicate numerical changes relative to Full: percentage points for
  accuracy/F1 and relative percentages for unrounded token counts.}
  \label{tab:core_ablation}
  \setlength{\tabcolsep}{5pt}
  \renewcommand{\arraystretch}{1.35}
  % Keep bold values at their regular-font width.
  \newlength{\ablationnumberwidth}
  \newcommand{\ablationbest}[2]{%
    \begingroup
    \settowidth{\ablationnumberwidth}{#1}%
    \setlength{\fboxsep}{0pt}%
    \resizebox{\ablationnumberwidth}{\height}{#2}%
    \endgroup}
  \resizebox{\textwidth}{!}{%
  \begin{tabular}{l|r@{\hspace{3pt}}lr@{\hspace{3pt}}l|r@{\hspace{3pt}}lr@{\hspace{3pt}}l}
    \toprule
    \multirow{2}{*}{\textbf{Variant}}
      & \multicolumn{4}{c|}{\textbf{MMLU-Pro}}
      & \multicolumn{4}{c}{\textbf{TAT-QA}} \\
    \cmidrule(lr){2-5}\cmidrule(lr){6-9}
      & \multicolumn{2}{c}{\textbf{Acc. (\%)} $\uparrow$}
      & \multicolumn{2}{c|}{\textbf{Tokens} $\downarrow$}
      & \multicolumn{2}{c}{\textbf{F1 (\%)} $\uparrow$}
      & \multicolumn{2}{c}{\textbf{Tokens} $\downarrow$} \\
    \midrule
    All-Atomic
      & 69.00 & {\scriptsize $\downarrow8.00$}
      & \ablationbest{$6.2\times10^{5}$}{$\boldsymbol{6.2\times10^{5}}$} & {\scriptsize $\downarrow44.76\%$}
      & 77.58 & {\scriptsize $\downarrow5.93$}
      & \ablationbest{$7.8\times10^{5}$}{$\boldsymbol{7.8\times10^{5}}$} & {\scriptsize $\downarrow11.57\%$} \\
    All-Group
      & \underline{71.00} & {\scriptsize $\downarrow6.00$}
      & $2.4\times10^{6}$ & {\scriptsize $\uparrow110.61\%$}
      & 78.59 & {\scriptsize $\downarrow4.92$}
      & $2.8\times10^{6}$ & {\scriptsize $\uparrow211.40\%$} \\
    Initialization only
      & 67.00 & {\scriptsize $\downarrow10.00$}
      & $1.3\times10^{6}$ & {\scriptsize $\uparrow15.52\%$}
      & 78.62 & {\scriptsize $\downarrow4.89$}
      & $1.3\times10^{6}$ & {\scriptsize $\uparrow49.30\%$} \\
    Final-reward only
      & 70.00 & {\scriptsize $\downarrow7.00$}
      & $1.3\times10^{6}$ & {\scriptsize $\uparrow18.03\%$}
      & \underline{82.99} & {\scriptsize $\downarrow0.52$}
      & $1.1\times10^{6}$ & {\scriptsize $\uparrow18.63\%$} \\
    \midrule
    \textbf{\methodname{} (Full)}
      & \ablationbest{77.00}{\textbf{77.00}} & {\scriptsize (ref.)}
      & \underline{$1.1\times10^{6}$} & {\scriptsize (ref.)}
      & \ablationbest{83.51}{\textbf{83.51}} & {\scriptsize (ref.)}
      & \underline{$8.9\times10^{5}$} & {\scriptsize (ref.)} \\
    \bottomrule
  \end{tabular}%
  }
\end{table*}

Table~\ref{tab:core_ablation} shows that Full achieves the highest performance
on both datasets. All-Atomic uses the fewest tokens but loses 8.00 accuracy
points on MMLU-Pro and 5.93 F1 points on TAT-QA; All-Group consumes
substantially more tokens while also underperforming Full. Mixed granularity
therefore improves performance over both fixed endpoints while avoiding
the cost of exclusively group-based execution. Full also improves on
Initialization only with fewer tokens, supporting learned construction.
Compared with Final-reward only, dense feedback yields a 7.00-point accuracy
gain on MMLU-Pro and a smaller 0.52-point F1 gain on TAT-QA, while reducing
tokens by 15.28\% and 15.70\%, respectively. These results support both local
granularity selection and intermediate construction feedback for learning
effective, economical organizations.

\subsection{Matched Optimization Comparison}
\label{sec:matched_optimization}

To answer \textbf{Q3}, we compare Search-then-SFT (SFT) and Dense-reward RL (RL)
from the same initialization under matched Qwen Flash API training budgets.
SFT alternates candidate search and verification with imitation of
the highest-scoring candidates with positive task scores; RL directly updates from
current-policy trajectories using dense feedback. At successive budget
stages, we evaluate both routes on the same 200 held-out MMLU-Pro queries
and 200 held-out TAT-QA queries, using accuracy and official F1, respectively
(Appendix~\ref{app:q3_protocol}).

\newpage
\begin{wrapfigure}{r}{0.4\textwidth}
  \vspace{-\intextsep}
  \centering
  \includegraphics[width=\linewidth]{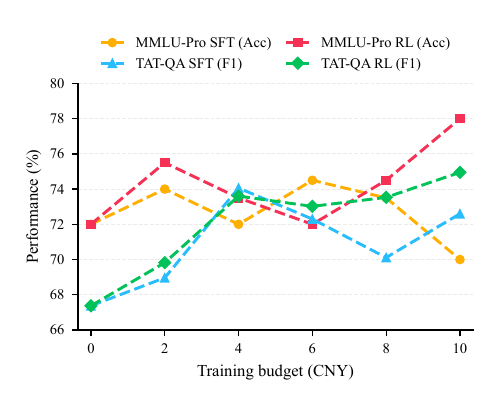}
  \caption{Performance versus Qwen Flash API training budget.}
  \label{fig:budget_performance}
\end{wrapfigure}

Figure~\ref{fig:budget_performance} shows that Dense-reward RL outperforms
Search-then-SFT at four of the five nonzero budget stages on each dataset.
At CNY~10, it reaches 78.00\% accuracy on MMLU-Pro and 74.95\% F1 on TAT-QA,
exceeding Search-then-SFT by 8.00 and 2.35 percentage points, respectively.
The advantage across most budget stages supports using construction-level
returns directly for policy updates, rather than first filtering candidates
into demonstrations, to obtain stronger policies within the available
API budget.

% Finish this wrapping region before the next subsection starts.
\par
\vspace{\dimexpr\value{WF@wrappedlines}\baselineskip\relax}
\WFclear

\subsection{Performance--Cost Trade-off}
\label{sec:performance_cost_tradeoff}

To answer \textbf{Q4}, we compare performance and total inference-token
consumption with alternative fixed and learned agent organizations on four
representative benchmarks, reusing the Q1 evaluations. We count input and
output tokens across all execution calls, including group-internal work
and final aggregation; training costs are accounted for separately.

\begin{figure}[!ht]
  \centering
  \includegraphics[width=\linewidth]{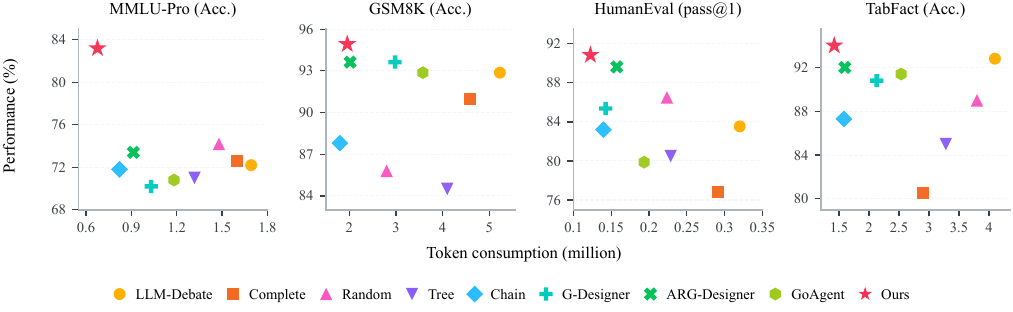}
  \caption{Performance--token trade-off across four datasets. The x-axis
  reports total inference-token consumption, and the y-axis reports task
  performance (accuracy or pass@1). Points toward the upper left indicate a
  more favorable trade-off.}
  \label{fig:performance_token_pareto}
\end{figure}

Figure~\ref{fig:performance_token_pareto} shows that \methodname{} lies on the
Pareto frontier on all four datasets: it has the highest performance throughout
and the lowest token use on MMLU-Pro, HumanEval, and TabFact. On GSM8K, Chain
uses fewer tokens but is 7.12 percentage points less accurate.
Thus, \methodname{} combines strong task performance with economical
inference. This trade-off is consistent with its mixed-granularity design,
which allows collaboration to be introduced locally while retaining
single-agent execution for other units.
\FloatBarrier

\section{Conclusion}

We presented \methodname{}, a unified perspective on task-conditioned MAS
design as mixed-granularity construction, where each role is locally realized
as an atomic agent or reusable group. Without a pre-collected successful
corpus, \methodname{} learns from current-policy trajectories using
potential-shaped returns and a regularized policy-gradient loss. Shaping
provides construction-level feedback while preserving the cumulative task
reward. Across four task families,
eight benchmarks, and 17 baselines, it ranks first on every benchmark. On four
representative benchmarks, it remains on the performance--token Pareto frontier
and uses the fewest inference tokens on three, supporting efficient
task-adaptive organization.

\bibliography{references}
\bibliographystyle{iclr2027/iclr2027_conference}

\appendix
\section{Detailed Method and Algorithms}
\label{app:implementation_training_details}

This section expands the construction policy, executable mixed-granularity agent graph, and training procedure of \methodname{} in Section~\ref{sec:methodology}. We first specify how the fixed library is executed, then describe the policy computation, potential-based rewards, and policy updates. The numerical settings below refer to the current documented configuration; diagnostic-specific settings are reported with their respective protocols.

\subsection{Role Realizations and Agent Graph Execution}
\label{app:policy_architecture}

A domain library supplies role profiles and their admissible atomic and group realizations. Each selected realization occupies one outer unit $u_i=(i,r_i,z_i)$, regardless of its internal size. An atomic realization uses one role executor. In the current group templates, three workers feed an internal aggregator, which emits the group's role-level result. For example, the calculation-auditor group combines a primary calculator, an independent recalculator, and a magnitude-and-unit checker with a calculation aggregator. Group membership, internal edges, role prompts, and tool permissions are fixed by the library; the policy chooses the outer roles, realizations, dependencies, and stopping position.

Outer edges point from existing units to newly appended units. During execution, groups expand into their fixed internal DAGs and agents run in dependency order. Each node request includes the public question, context, answer-format requirements, and a \texttt{predecessors} JSON array. Its structured message objects retain predecessor identifiers, roles, granularities, status, and payload fields such as analysis, answers, and tool observations. The request is serialized as JSON text for the execution LLM. Message projection and length budgets bound the supplied content.

All three workers in a group receive the results of every external direct predecessor. The internal aggregator receives the three workers as its direct predecessors, together with the public task, answer requirements, and role description. External predecessor messages are not additionally routed directly to this aggregator. The group exposes only the aggregator's result. The fixed outer summarizer receives the original question, output requirements, and results from all sinks of the expanded graph. This execution rule applies to nonempty partial graphs used for probes as well as completed graphs. An empty graph instead invokes the base LLM with the question, available answer options, and answer requirement, as defined in Section~\ref{sec:method_graph}.

The construction policy observes the public query and graph structure. Executor responses, probe scores, and gold answers do not enter its decision context. At inference, construction is completed before the organization is executed. Tool requests and their results are handled by the execution layer; training updates neither the tools nor the execution LLM. The current execution configuration uses temperature zero, disabled thinking, \texttt{json\_object} output, and a 2,048-token output limit. Core response content includes analysis and an answer, with field validity handled by the execution parser. The selected execution backbone and full prompt templates belong to the experiment-specific configuration.

\subsection{Task-Conditioned Construction Policy}
\label{app:construction_policy}

\paragraph{Frozen semantic features.}

The \texttt{sentence-transformers/all-MiniLM-L6-v2} encoder produces 384-dimensional semantic vectors. The public task view separates the question and options, public context, and metadata such as dataset and role/tool information. Long fields are chunked at the tokenizer limit. Each chunk is L2-normalized; chunk vectors are averaged and normalized again to obtain a field vector. With missing fields omitted, the task embedding is

\begin{equation}
\mathbf q=\operatorname{Normalize}
 \left(0.5\mathbf e_{\mathrm{question}}
       +0.4\mathbf e_{\mathrm{context}}
       +0.1\mathbf e_{\mathrm{metadata}}\right).
\label{app:eq:task_embedding}
\end{equation}

Answer-format instructions are supplied to the executors separately. Task and role-realization embeddings are cached and frozen during training.

\paragraph{Graph features and recurrence.}

Each outer unit has a 256-dimensional feature vector

\begin{equation}
\mathbf x_i=W_s\mathbf e_i+\mathbf e_{z_i}
 +\mathbf e_{\mathrm{pos}(i)}
 +\mathbf e_{\deg^-(i)}+\mathbf e_{\deg^+(i)},
\label{app:eq:node_features}
\end{equation}

where $\mathbf e_i$ is its frozen realization embedding. The semantic projection, granularity embedding, construction-position embedding, and outer-degree embeddings are trainable. Position, indegree, and outdegree each use a separate $N_{\max}\times256$ embedding table, currently $3\times256$. Construction positions are indexed by 0, 1, and 2; degrees are measured on the outer agent graph. Unsupported indices raise an error rather than being clamped. A shared Edge-aware GRUCell processes units in construction order:

\begin{equation}
\begin{aligned}
\mathbf p_i&=\operatorname{Mean}_{j\to i}\mathbf h_j, &
 \mathbf v_i&=W_x\mathbf x_i+W_p\mathbf p_i+W_q\mathbf q,\\
 \mathbf u_i&=\operatorname{GRUCell}
   (\operatorname{Dropout}(\mathbf v_i),\mathbf u_{i-1}), &
 \mathbf h_i&=\operatorname{LayerNorm}(\mathbf u_i).
\end{aligned}
\label{app:eq:edge_gru}
\end{equation}

An empty predecessor set gives $\mathbf p_i=\mathbf0$, and $\mathbf u_0=\mathbf0$. The recurrent state represents construction history; predecessor aggregation represents actual outer dependencies. Both belong to the policy's structural encoding. Agent-to-agent execution messages continue to follow the selected edges. The current graph is re-encoded at every construction decision, including its updated degrees.

\paragraph{Pooling and decision context.}

Task-conditioned attention pools the node representations:

\begin{equation}
a_i=\operatorname{softmax}_i
 \left(\frac{(W_k\mathbf h_i)^\top W_a\mathbf q}{\sqrt{256}}\right),
 \qquad \mathbf g_{\mathrm{pool}}=\sum_i a_i\mathbf h_i.
\label{app:eq:graph_pooling}
\end{equation}

For an empty graph, a learned 256-dimensional vector replaces $\mathbf g_{\mathrm{pool}}$. Concatenating the pooled vector with $\mathbf q$ and applying a linear projection and LayerNorm gives a 256-dimensional graph vector $\mathbf g_t$. The action context is

\begin{equation}
\mathbf c_t=\operatorname{LayerNorm}\left(
 \operatorname{GELU}\left(W_c
 [\mathbf g_t;\mathbf q;N_t/N_{\max};1-N_t/N_{\max}]+\mathbf b_c\right)
 \right),
\label{app:eq:decision_context}
\end{equation}

where $N_t$ is the current outer-unit count and $W_c$ maps 642 inputs to 256 outputs.

\paragraph{Hierarchical actions.}

A shared role-scoring MLP receives $\mathbf c_t$ and each role's semantic embedding; a separate MLP scores \textsc{stop} from $\mathbf c_t$. These scores enter one masked categorical distribution. Conditional on the sampled role, the granularity MLP receives the context, role semantics, and atomic and group semantics to score the admissible realizations. Conditional on both choices, the connection head forms a query from their semantics and context, and keys from existing node states. Each edge score combines a query--key dot product and an interaction MLP over the query, node state, and their coordinate-wise product. Legal predecessor indicators are sampled as independent Bernoulli variables. The current head dimensions are:

\begin{center}
\small
\begin{tabular}{ll}
\toprule
Module & Dimensions and transformation \\
\midrule
Role MLP & $640\to256\to1$, intermediate GELU; shared across roles \\
\textsc{stop} MLP & $256\to256\to1$, intermediate GELU \\
Granularity MLP & $1408\to256\to2$, intermediate GELU \\
Connection query & $1024\to256$, linear with bias \\
Connection key & $256\to256$, linear without bias \\
Connection interaction MLP & $768\to256\to1$, intermediate GELU \\
\bottomrule
\end{tabular}
\end{center}

For connection query $\mathbf v_{r,z}$ and key $\mathbf k_j$ obtained from node state $\mathbf h_j$, the edge score is

\begin{equation}
\ell_{r,z,j}=\mathbf v_{r,z}^{\top}\mathbf k_j+
\operatorname{MLP}([\mathbf v_{r,z};\mathbf h_j;\mathbf v_{r,z}\odot\mathbf h_j]).
\label{app:eq:connection_score}
\end{equation}

The connection dot product is unscaled, and its interaction term uses the node state $\mathbf h_j$. Task-conditioned graph pooling uses the $1/\sqrt{256}$ scaling in Equation~\ref{app:eq:graph_pooling}. The decision heads contain no dropout.

Writing $p_{j,t}$ for the probability of predecessor edge $j$, the sampled non-\textsc{stop} action has log-probability

\begin{equation}
\begin{aligned}
\log\pi_\theta(a_t\mid s_t)
 ={}&\log\pi_\theta(r_t\mid s_t)
 +\log\pi_\theta(z_t\mid r_t,s_t)\\
 &+\sum_{j\in\mathcal V_t}
 \left[b_{j,t}\log p_{j,t}+(1-b_{j,t})\log(1-p_{j,t})\right].
\end{aligned}
\label{app:eq:action_logprob}
\end{equation}

Both selected and absent legal edges contribute. A sampled \textsc{stop} uses only its role-layer probability. The first unit has no existing predecessor, so its edge term is an empty sum. Masks restrict choices to available roles and realizations, compatible interfaces, and the expanded-depth bound. Predecessors must be existing units and edges follow construction order, with self-loops and duplicate predecessor entries excluded. Zero-predecessor additions remain legal after the first unit, allowing independent branches. Repeated roles are also legal and contribute to the role-repetition component of the shaping potential. There is no separate maximum-indegree parameter.

Expanded depth counts nodes on the longest path, with root depth one. It includes group workers and internal aggregators and excludes the outer summarizer. An independent atomic unit has depth one, an independent group has depth two, and two serial groups have depth four. \textsc{stop} is always legal. At the outer-unit limit, all non-\textsc{stop} roles are masked, leaving a probability-one \textsc{stop} action. This action remains an explicit position in the trajectory.

The documented policy uses one shared GRUCell with hidden size 256, dropout 0.1 on each GRU input projection sum during training, inference policy sampling temperature 1.0, a maximum of three outer units, and maximum expanded depth four. Evaluation disables dropout and retains probabilistic graph sampling. The policy starts from random initialization without \textsc{stop} warmup. Semantic projections, structural embeddings, the GRU, attention pooling, context projection, and action heads are trainable; MiniLM and the execution LLM remain frozen.

\subsection{Potential-Based Rewards and Policy Optimization}
\label{app:reward_specification}

\paragraph{Probe utility and evaluation reuse.}

Four probe questions are selected without replacement from the remaining training pool using a reproducible, seed-dependent pseudorandom hash ordering, and are held fixed throughout each trajectory group as the probe set $P$. The current training query is excluded. Candidate hash keys depend on the experiment seed, current query ID, update index, and data-source version; the first four candidates in this ordering are selected. Probe questions may recur across updates. Each partial graph is executed with its selected roles and edges on the probe questions, providing a shared auxiliary measure of graph utility across these training queries. The terminal score separately evaluates the completed graph on the conditioning query. For score $S(\mathcal G,p)\in[0,1]$, define

\begin{equation}
U_P(\mathcal G)=
 \frac{\kappa u_0+\sum_{p\in P}S(\mathcal G,p)}{\kappa+|P|}.
\label{app:eq:probe_utility}
\end{equation}

Parent--child differences cancel the shared prior:

\begin{equation}
\Delta U_P=
 \frac{\sum_{p\in P}[S(\mathcal G',p)-S(\mathcal G,p)]}{\kappa+|P|}.
\label{eq:marginal_utility_reward}
\end{equation}

The evaluation ledger stores reusable graph--query execution results. A missing graph--probe pair is executed and recorded; an available reusable record is read directly. Probe membership stays fixed independently of cache availability. Terminal and probe executions populate the ledger according to their execution paths. These records support reward computation and are not inputs to the construction policy.

\paragraph{Structural potential.}

Let $N_{\mathrm{exp}}$ and $E_{\mathrm{exp}}$ count execution nodes and edges after group expansion. With $[x]_+=\max(0,x)$, define

\begin{equation}
\begin{aligned}
C(\mathcal G)&=w_n[N_{\mathrm{exp}}(\mathcal G)-b_n]_+
              +w_e[E_{\mathrm{exp}}(\mathcal G)-b_e]_+,
 \\
 D(\mathcal G)&=\sum_{r\in\mathcal R}[m_r(\mathcal G)-1]_+,
\end{aligned}
\label{eq:structural_penalties}
\end{equation}

where $m_r$ counts outer units assigned role $r$ across both realizations. For every addition, the accumulated potential changes by

\begin{equation}
d_t=\alpha\Delta U_P-\beta\Delta C-\gamma_R\Delta D,
 \qquad \Phi_{t+1}=\Phi_t+d_t,\quad\Phi_0=0.
\label{app:eq:potential_increment}
\end{equation}

With fixed probes and consistent recorded evaluations, this accumulation implements the initial-graph-relative potential defined in Section~\ref{sec:method_optimization}.

\paragraph{Rewards and termination.}

An ordinary addition receives shaping reward $F_t=\eta\Phi_{t+1}-\Phi_t$. Every trajectory ends with an explicit \textsc{stop} action, including when the outer-unit limit makes \textsc{stop} the only legal choice. At this position, the graph receives its task score and the potential is fully settled to zero. The trajectory stores addition rewards in \texttt{transitions}, all addition and \textsc{stop} decisions in \texttt{actions}, and terminal scoring and settlement separately as \texttt{terminal\_reward} and \texttt{terminal\_shaping\_reward}. The settlement equals the negative pre-\textsc{stop} potential and is retained even when negative.

Let $T_k$ be the \textsc{stop} action index. Return computation combines the two terminal fields at this position and then proceeds backward:

\begin{equation}
\begin{aligned}
G_{k,T_k}&=r_{\mathrm{terminal},k}+r_{\mathrm{terminal\_shaping},k}
=S(\mathcal G_k,q)-\Phi_{k,T_k},\\
G_{k,t}&=r^{\mathrm{dense}}_{k,t}+\eta G_{k,t+1}\quad(t<T_k).
\end{aligned}
\label{app:eq:terminal_returns}
\end{equation}

The terminal terms are kept separate from the final addition's stored dense reward. Immediate \textsc{stop} has zero initial potential and receives the direct-answer score. A horizon-forced \textsc{stop} has log-probability zero but still occupies an action position for return normalization and loss averaging.

Terminal scores use accuracy for ordinary answer tasks, official F1 for TAT-QA, and single-sample test success for coding, reported as pass@1. The current call and token reward weights are zero. Reward parameters are $\alpha=1$, $\beta=0.02$, $\gamma_R=0.05$, $\eta=1$, $\kappa=1$, $u_0=0.5$, $w_n=w_e=1$, and $b_n=b_e=6$. The free allowances apply to expanded nodes and edges. Structural terms redistribute intermediate credit; full terminal settlement leaves the trajectory's total reward equal to its task score.

\paragraph{Position-aligned returns and advantages.}
\label{app:grpo_objective}

For the current configuration, $K=2$ trajectories are sampled for each query. Let $L_k=T_k+1$ count recorded action positions in trajectory $k$, including its final \textsc{stop} even when forced by the horizon. After incorporating terminal scoring and settlement, compute discounted returns backward with the same $\eta$ used for shaping. For $\eta=1$, the return at an actual action position is

\begin{equation}
G_{k,t}=S(\mathcal G_k,q)-\Phi_{k,t},
\label{eq:step_credit}
\end{equation}

where $\mathcal G_k$ is the final graph and $\Phi_{k,t}$ is the potential before that action. For $I_t=\{k:t<L_k\}$ and $n_t=|I_t|$, compute

\begin{equation}
\begin{aligned}
\mu_t&=\frac1{n_t}\sum_{k\in I_t}G_{k,t}, &
 \sigma_t&=\left(\frac1{n_t}\sum_{k\in I_t}(G_{k,t}-\mu_t)^2\right)^{1/2},
 \\
 A_{k,t}&=\frac{G_{k,t}-\mu_t}{\sigma_t+\epsilon}, & \epsilon&=10^{-8}.
\end{aligned}
\label{app:eq:position_advantage}
\end{equation}

The standard deviation uses the population denominator. For numerical stability, the implementation divides returns at each position by their maximum absolute value $m_t=\max_{k\in I_t}|G_{k,t}|$ and scales epsilon to $\epsilon/m_t$. When $m_t>0$, this gives Equation~\ref{app:eq:position_advantage} in exact arithmetic. All-zero returns, zero standard deviation, and singleton positions yield zero advantage. Short trajectories supply no padding values. Alignment follows action index, so a \textsc{stop} may be compared with an addition at the same position in another trajectory. Advantages are detached before policy optimization.

\paragraph{Hierarchical KL and entropy.}

The reference policy is a deep copy made before the first RL update, with all parameters frozen and evaluation mode enabled throughout training. At each visited state, both policies use the same legal action support. For $X\in\{\mathcal K,\mathcal H\}$ denoting KL or entropy, respectively,

\begin{equation}
X(s)=X_r(s)+\sum_{r\ne\mathrm{STOP}}\pi_\theta(r\mid s)
 \left[X_z(s,r)+\sum_z\pi_\theta(z\mid r,s)
                    \sum_{j\in\mathcal V(s,r,z)}X_{e_j}(s,r,z)\right].
\label{app:eq:hierarchical_regularizer}
\end{equation}

For $\mathcal K$, the terms are categorical or Bernoulli divergences in the current-to-reference direction; for $\mathcal H$, they are current-policy entropies. Conditional terms are weighted by current-policy probabilities. The sums cover all legal branches at the visited state, including when the sampled action is \textsc{stop}. Edge terms are summed over legal predecessors, and logarithms are natural.

Sampled action log-probabilities retain their computation graphs. For a non-\textsc{stop} action, KL and entropy reuse the stored sampling output and hence the same dropout realization. \textsc{stop} decoding skips conditional heads; to obtain the full hierarchical distribution, the implementation restores the pre-sampling RNG state and repeats the forward pass. It checks that the reproduced role logits match the sampled logits elementwise. Thus the current-policy regularizers share the sampling dropout realization, while the reference policy always has dropout disabled.

With $M=\sum_k L_k$, including every final \textsc{stop} position, the action-averaged loss is

\begin{equation}
\mathcal L=\frac1M\sum_k\sum_{t=0}^{L_k-1}
 \left[-\operatorname{sg}(A_{k,t})\log\pi_\theta(a_{k,t}\mid s_{k,t})
 +\beta_{\mathrm{KL}}\mathcal K(s_{k,t})
 -\beta_H\mathcal H(s_{k,t})\right].
\label{app:eq:grouped_policy_objective}
\end{equation}

Each recorded action position, including a probability-one \textsc{stop} at the horizon, receives equal weight. KL and entropy enter the loss separately from the rewards and can contribute gradients at zero-advantage positions. The optimizer is Adam with fixed learning rate $10^{-4}$ and otherwise default parameters, without gradient clipping or a learning-rate schedule; $\beta_{\mathrm{KL}}=0.01$, and $\beta_H=0.001$. Only the construction policy parameters listed above are updated.

\subsection{Training and Inference Procedures}
\label{app:training_inference_algorithms}

The procedures below specify the current undiscounted setting. Ledger lookups reuse existing evaluations; each newly executed graph--query pair incurs execution cost, including group members, tool-related calls when applicable, and final summarization. Probe execution is confined to training.

\paragraph{Procedure A1: One training update.}

\begin{enumerate}
\item \textbf{Inputs:} training query $q$, its scorer, fixed library, current policy $\pi_\theta$, frozen RL-initial reference $\pi_{\mathrm{ref}}$, training query pool, and evaluation ledger. Select a shared probe set $P$ excluding $q$ and encode the public task.
\item \textbf{Rollouts:} for each of $K$ trajectories, initialize $\mathcal G\leftarrow\mathcal G_0$ and $\Phi\leftarrow0$. Until termination, encode the current graph and sample a masked role-or-\textsc{stop} action. Record the visited state and action log-probability with its computation graph. Retain the sampling output, or replay the pre-sampling RNG state for \textsc{stop}, to obtain the full current-policy distribution with the same dropout realization.
\item \textbf{Addition:} if a role is selected, sample its realization and legal predecessor indicators, completing the recorded action probability. Append the unit to obtain $\mathcal G'$. Retrieve or execute parent and child probe evaluations; compute $\Delta U_P$, $\Delta C$, and $\Delta D$. Record reward $d=\alpha\Delta U_P-\beta\Delta C-\gamma_R\Delta D$ and set $(\mathcal G,\Phi)\leftarrow(\mathcal G',\Phi+d)$. Return to the role decision; at the outer-unit horizon, mask all non-\textsc{stop} roles and record the resulting probability-one \textsc{stop}.
\item \textbf{Termination:} on \textsc{stop}, score the final graph on $q$ and store \texttt{terminal\_reward} $=S(\mathcal G,q)$ and \texttt{terminal\_shaping\_reward} $=-\Phi$. Keep this \textsc{stop} in the action sequence. Set its return to the sum of the terminal fields; leave previously stored addition rewards unchanged.
\item \textbf{Update:} compute returns backward for every trajectory, normalize by actual action position using Equation~\ref{app:eq:position_advantage}, and detach the advantages. Compute hierarchical KL and entropy on the recorded states using the retained or replayed current-policy outputs and the frozen reference in evaluation mode. Include the \textsc{stop} positions in normalization and action averaging. Minimize Equation~\ref{app:eq:grouped_policy_objective} with Adam. \textbf{Outputs:} updated construction-policy parameters and ledger.
\end{enumerate}
\paragraph{Procedure A2: Inference.}

\begin{enumerate}
\item \textbf{Inputs:} public query $q$, fixed library, trained policy, and frozen execution LLM. Disable policy dropout and initialize the empty graph.
\item Re-encode the graph and sample the role-or-\textsc{stop} decision. On a role decision, sample its realization and legal predecessor indicators, append the unit, and repeat. At the outer-unit horizon, only \textsc{stop} remains legal. Terminate when \textsc{stop} is selected.
\item Execute the completed agent graph and summarize the expanded graph's sink results together with the public task and output requirements, or use the direct-answer fallback for an empty graph. \textbf{Output:} final answer.
\end{enumerate}

\section{Proof of Return Preservation}
\label{app:shaping_proof}

This section proves Proposition 1 in Section~\ref{sec:shaping_theory}, following potential-based reward shaping~\citep{ng1999policy}, and derives the return-to-go used in Appendix~\ref{app:implementation_training_details}. The construction episode includes its final \textsc{stop} action, also when the outer-unit limit leaves \textsc{stop} as the only legal choice.

\subsection{Episode and Potential Conventions}

Fix a query and a probe set shared throughout its trajectory group. Let an episode contain $T$ action transitions, indexed by $t=0,\ldots,T-1$, with \textsc{stop} at index $T-1$. State $s_t$ is the state before action $t$, and $s_T$ is the terminal state after \textsc{stop}. For $N$ additions, $T=N+1$; immediate \textsc{stop} gives $T=1$. The horizon limits additions while retaining the final \textsc{stop} position.

During construction, the potential is defined relative to the initial graph using the fixed probes, structural complexity, and role repetition specified in Appendix~\ref{app:implementation_training_details}. Evaluation records and the accumulated potential can be included in the training state to make this quantity well-defined along the realized episode. The policy continues to observe only the public task and graph features. We set

\begin{equation}
\Phi_0=0,\qquad \Phi_T=0.
\label{app:eq:proof_1}
\end{equation}

Here $\Phi_{T-1}$ is the potential of the completed graph before \textsc{stop}. Its settlement sets the terminal-state potential to zero while retaining the graph's task score. Let $S$ denote that score and define the base rewards by

\begin{equation}
r_t^{\mathrm{task}}=
\begin{cases}
0,&0\le t<T-1,\\
S,&t=T-1.
\end{cases}
\label{app:eq:proof_2}
\end{equation}

Using the same discount $\eta$ for shaping and return computation, each shaped reward is

\begin{equation}
r'_t=r_t^{\mathrm{task}}+\eta\Phi_{t+1}-\Phi_t.
\label{app:eq:proof_3}
\end{equation}

In particular, the \textsc{stop} reward is $S-\Phi_{T-1}$. The implementation stores the score and settlement separately and combines them at the \textsc{stop} position when computing returns.

\subsection{Proof of Proposition 1}

\textbf{Proposition 1 (Return preservation).} For a finite episode with the endpoint conditions in Equation~\ref{app:eq:proof_1}, the shaped and base discounted returns coincide:

\begin{equation}
\sum_{t=0}^{T-1}\eta^t r'_t
=\sum_{t=0}^{T-1}\eta^t r_t^{\mathrm{task}}.
\label{app:eq:proof_4}
\end{equation}

\textbf{Proof.} Substitute Equation~\ref{app:eq:proof_3} and collect the potential terms:

\begin{equation}
\begin{aligned}
\sum_{t=0}^{T-1}\eta^t r'_t
&=\sum_{t=0}^{T-1}\eta^t r_t^{\mathrm{task}}
 +\sum_{t=0}^{T-1}\eta^{t+1}\Phi_{t+1}
 -\sum_{t=0}^{T-1}\eta^t\Phi_t\\
&=\sum_{t=0}^{T-1}\eta^t r_t^{\mathrm{task}}
 -\Phi_0+\eta^T\Phi_T\\
&=\sum_{t=0}^{T-1}\eta^t r_t^{\mathrm{task}}.
\end{aligned}
\label{app:eq:proof_5}
\end{equation}

Every interior potential cancels with its counterpart from the adjacent transition, and both endpoint terms vanish. The identity holds for each realized episode, including immediate \textsc{stop} and the probability-one \textsc{stop} at the unit horizon. Taking expectations over task queries, probe selection, policy samples, and execution outcomes therefore gives the same expected return for every policy under the two reward definitions. Their maximizing policy sets coincide, including within the policy class used by \methodname{}. $\square$

For a general discount, the common return is $\eta^{T-1}S$. In the implemented setting $\eta=1$, it is exactly $S$, so potential shaping preserves the expected terminal task-score objective in the main text. Utility, structural complexity, and role repetition determine the intermediate allocation of reward, with their net contribution settled at \textsc{stop}.

\subsection{Return-to-Go and Position-Aligned Credit}

Applying the same cancellation from an arbitrary action position $t$ gives

\begin{equation}
\begin{aligned}
G'_t
&=\sum_{j=t}^{T-1}\eta^{j-t}r'_j\\
&=G_t^{\mathrm{task}}-\Phi_t+\eta^{T-t}\Phi_T\\
&=\eta^{T-1-t}S-\Phi_t.
\end{aligned}
\label{app:eq:proof_6}
\end{equation}

Thus, for the current undiscounted configuration,

\begin{equation}
G'_t=S-\Phi_t.
\label{app:eq:proof_7}
\end{equation}

The first action has return $S$ because its potential is zero. At later positions, the potential provides a state-dependent reference for the partial organization built before the action. For two trajectories with the same terminal score, $G'_{i,t}-G'_{j,t}=-(\Phi_{i,t}-\Phi_{j,t})$: their returns differ according to their pre-action graph potentials. Position-aligned normalization converts these state-relative returns into advantages that weight the sampled construction actions. At the final \textsc{stop} position, Equation~\ref{app:eq:proof_7} gives precisely the score plus terminal settlement, including for a horizon-forced \textsc{stop}.

Return preservation concerns the environment objective. The learning procedure subsequently normalizes returns within each action position and applies hierarchical KL and entropy regularization as specified in Appendix~\ref{app:implementation_training_details}. Keeping these operations separate makes explicit how shaping preserves the task objective while supplying the returns used by the implemented optimizer.

\section{Experimental Configuration and Protocols}
\label{app:experimental_details}

This section separates shared resources from experiment-specific data allocations
and run settings. Appendix~\ref{app:implementation_training_details} defines
the method and algorithms; Appendix~\ref{app:extended_empirical_statistics}
contains the diagnostic protocols. Source split names describe dataset
provenance, whereas training, probe, development, and evaluation describe an
example's role in a particular experiment.

\subsection{Experiment Overview}
\label{app:experiment_overview}

\begin{table}[!ht]
\centering
\small
\setlength{\tabcolsep}{4pt}
\begin{tabular}{@{}p{0.13\linewidth}p{0.40\linewidth}p{0.34\linewidth}@{}}
\toprule
Experiment & Datasets & Protocol \\
\midrule
D1 & MMLU-Pro; TAT-QA & Appendix~\ref{app:diagnostic_d1} \\
D2 & Analytical configuration coverage & Appendix~\ref{app:extended_empirical_statistics} \\
D3 & MMLU-Pro; GSM8K; HumanEval & Appendix~\ref{app:diagnostic_d3} \\
Q1 / Q4 & Eight main benchmarks / four efficiency benchmarks & Appendix~\ref{app:q1_q4_protocol} \\
Q2 & MMLU-Pro; TAT-QA & Appendix~\ref{app:q2_protocol} \\
Q3 & MMLU-Pro; TAT-QA & Appendix~\ref{app:q3_protocol} \\
\bottomrule
\end{tabular}
\caption{Experiment-to-protocol index. Data allocations and settings apply only
to the experiments explicitly identified in each protocol.}
\label{app:tab:experiment_index}
\end{table}
\FloatBarrier

\subsection{Shared Resources and Evaluation Conventions}
\label{app:datasets_scoring}

\paragraph{Benchmarks and data roles.}
The main evaluation covers eight benchmarks across four task families.
The source inventory below describes the dataset collections, not the samples
used by every experiment. Q1 allocations are listed in
Appendix~\ref{app:q1_q4_protocol}; Q2, Q3, D1, and D3 have separate protocols.
Scores are expressed as percentages. TAT-QA uses official F1, coding tasks
use pass@1, and the remaining main tasks use accuracy. D1's TAT-QA diagnostic
uses its explicitly specified strict-answer scoring.
Dataset versions and membership lists must identify the actual snapshots
and examples used in each run.

\begin{table}[!ht]
\centering
\small
\setlength{\tabcolsep}{3pt}
\renewcommand{\arraystretch}{1.12}
\caption{Source dataset inventory for the main-table runs.}
\label{app:tab:source_datasets}
\begin{tabular}{@{}p{0.2\linewidth}p{0.22\linewidth}p{0.28\linewidth}p{0.2\linewidth}@{}}
\toprule
\raggedright Benchmark & \raggedright Source train & \raggedright Source validation/development & \raggedright Source test \tabularnewline
\midrule
\raggedright MMLU-Pro & \raggedright None & \raggedright 70 & \raggedright 12,032 \tabularnewline
\raggedright StrategyQA & \raggedright 2,061 under the official-code split of 2,290 public training questions & \raggedright 229 under that split & \raggedright 490 hidden examples \tabularnewline
\raggedright AQuA-RAT & \raggedright 97,467 & \raggedright 254 & \raggedright 254 \tabularnewline
\raggedright GSM8K & \raggedright 7,473 & \raggedright None & \raggedright 1,319 \tabularnewline
\raggedright HumanEval & \raggedright None & \raggedright None & \raggedright 164 \tabularnewline
\raggedright LiveCodeBench & \raggedright No standard training split & \raggedright No standard development split & \raggedright 1,055 cumulative examples in \texttt{release\_\allowbreak{}v6} \tabularnewline
\raggedright TAT-QA & \raggedright 13,215 & \raggedright 1,668 & \raggedright 1,669 \tabularnewline
\raggedright TabFact & \raggedright 92,283 & \raggedright 12,792 & \raggedright 12,779 \tabularnewline
\bottomrule
\end{tabular}
\end{table}
\FloatBarrier

\paragraph{Answer extraction and scoring.}

The scoring metric for each main-evaluation benchmark follows Table~\ref{app:tab:dataset_allocation}. For training, rewards use the corresponding score on a scale from zero to one; TAT-QA uses official F1, with EM recorded separately in the documented reward configuration. Coding rewards are single-sample test-success indicators.

The D1 diagnostic uses a separate strict-answer protocol for TAT-QA. Its accuracy measurements and the official F1 scores in the main evaluation and component ablations are reported under their respective protocols. Diagnostic sampling and scoring details belong to Appendix~\ref{app:extended_empirical_statistics}.

The model produces a JSON object containing \texttt{analysis} and \texttt{answer}. Only \texttt{answer} is scored; the evaluator does not infer or recover answers from the analysis text. The executor deterministically adds the internal \texttt{FINAL:} wrapper required by the scoring interface, without an additional model call. A null or empty final answer, or failure to parse the final answer, receives zero credit. TAT-QA uses the official \texttt{NExTplusplus/TAT-QA} scorer pinned to commit \texttt{870accc4\allowbreak{}1953dcde\allowbreak{}885aabeb\allowbreak{}963d94aa\allowbreak{}bdc0fbc3}. Dataset answer contracts are specified in Appendix~\ref{app:role_protocols}.

MMLU-Pro, AQuA, StrategyQA, TabFact, and GSM8K use binary scoring: a matching final answer receives one and any mismatch or parsing failure receives zero. Leading and trailing whitespace is removed before comparison. For MMLU-Pro and AQuA, the prediction must be a single English letter and is matched case-insensitively against the reference option label. MMLU-Pro integer reference indices are mapped as $0\mapsto A$, $1\mapsto B$, and so on; AQuA uses its \texttt{correct} field. Reference option labels are uppercased.

For StrategyQA and TabFact, predictions must be case-insensitive \texttt{true} or \texttt{false} strings and are compared as Boolean values. StrategyQA Boolean references are retained, with reference \texttt{yes}/\texttt{no} strings mapped to Boolean values. TabFact reference labels \texttt{1}/\texttt{entailed} map to true and \texttt{0}/\texttt{refuted} to false.

For GSM8K, the reference is the trimmed text following the last \texttt{\#\#\#\#} delimiter. Predictions must match \verb|-?[0-9.,]+| in full. After removing commas from both strings, scoring uses exact string equality. Thus, \texttt{1,000} matches \texttt{1000}, while \texttt{42.0} and \texttt{042} do not match \texttt{42}; numerical tolerance is not applied.

\paragraph{Validation, recovery, and failure handling.}

The \texttt{analysis}/\texttt{answer} object permits additional fields while requiring the mandatory fields and their correct types. Invalid JSON, duplicate keys, missing required fields, and type errors are recorded as format failures. TAT-QA's nested \texttt{values}/\texttt{scale} object remains strictly validated.

An intermediate node that fails semantic or contextual validation may receive one repair call. Format errors do not uniformly trigger repair. When an intermediate response is truncated at the 2,048-token output limit, one additional attempt may request a concise, complete answer. Direct-answer and outer-finalizer nodes do not receive these format-repair or truncation-recovery calls.

Network exceptions, timeouts, HTTP 429 responses, and HTTP 5xx responses allow at most four total attempts, with backoff intervals of 1, 2, and 4 seconds. SDK-level retries are disabled.

Node failures are recorded individually. Other graph nodes and the outer finalizer may still produce a valid answer, which is scored normally. An internal failure event therefore does not by itself set the example's score to zero. If no valid final answer is obtained, the score is zero; failed examples are retained in the evaluation denominator.

\paragraph{Code evaluation and pass@1.}

HumanEval and LiveCodeBench-v6 use a custom evaluation entry point that executes Python programs in a Bubblewrap sandbox inside a local Lima Linux virtual machine. For HumanEval, the evaluator retains necessary imports and helper definitions from the public prompt, removes the placeholder definition of the target function, loads the final program, and executes the supplied test code through \texttt{check(candidate)} using the task's \texttt{entry\_point}. Success requires all assertions to pass. For LiveCodeBench-v6, evaluation combines public and private test cases and supports both standard-input/output and function-call modes. Standard output is compared after stripping leading and trailing whitespace from each line, with numerical lines compared exactly using \texttt{Decimal}; function return values are compared using Python value equality. All test cases must pass.

Candidate execution is limited to 2 seconds of CPU time and 3 seconds of wall-clock time, with sandbox startup measured separately. The infrastructure has an overall 30-second wall-clock limit. Memory and output limits are 256 MiB and 64 KiB, respectively. For LiveCodeBench-v6, these execution limits are set when each test case starts, and CPU and execution wall-clock times are also accumulated across all cases and checked against the same 2-second and 3-second totals.

Pass@1 is the mean binary success score of one final program from one complete system run per question. Internal candidates, tool calls, and public-test executions are part of that run; hidden-test outcomes are used only to score the final submission. Syntax errors, runtime errors, failed tests, execution timeouts, and unparseable final answers receive zero credit. Sandbox unavailability and infrastructure timeouts also receive zero credit, with distinct execution statuses retained to identify their causes.

\subsection{Overall Performance and Inference Efficiency (Q1 / Q4)}
\label{app:q1_q4_protocol}

\paragraph{Q1 data allocation.}
Table~\ref{app:tab:dataset_allocation} applies to the overall-performance
comparison in Table~\ref{tab:main_results}. It does not specify the data pools
for the component, optimization, or diagnostic experiments.

\begin{table}[!ht]
\centering
\small
\setlength{\tabcolsep}{3pt}
\renewcommand{\arraystretch}{1.12}
\caption{Data allocation for Q1 overall performance.}
\label{app:tab:dataset_allocation}
\begin{tabular}{@{}p{0.17\linewidth}p{0.12\linewidth}p{0.35\linewidth}p{0.26\linewidth}@{}}
\toprule
\raggedright Benchmark & \raggedright Metric & \raggedright Study training examples & \raggedright Study test examples \tabularnewline
\midrule
\raggedright MMLU-Pro & \raggedright Accuracy & \raggedright 100 from source test & \raggedright 500 from source test \tabularnewline
\raggedright StrategyQA & \raggedright Accuracy & \raggedright 100 from the pooled source dev + train partitions & \raggedright 500 from the pooled source dev + train partitions \tabularnewline
\raggedright AQuA-RAT & \raggedright Accuracy & \raggedright 100 from source train & \raggedright All 254 source test examples \tabularnewline
\raggedright GSM8K & \raggedright Accuracy & \raggedright 100 from source train & \raggedright All 1,319 source test examples \tabularnewline
\raggedright HumanEval & \raggedright pass@1 & \raggedright The same 100 LiveCodeBench training examples & \raggedright All 164 source test examples \tabularnewline
\raggedright LiveCodeBench-v6 & \raggedright pass@1 & \raggedright 100 sampled from the remainder of \texttt{release\_\allowbreak{}v6}, excluding the v6 test subset & \raggedright All 175 examples with \texttt{version\_\allowbreak{}tag="v6"} \tabularnewline
\raggedright TAT-QA & \raggedright Official F1 & \raggedright 100 from source train & \raggedright All 1,669 source test examples \tabularnewline
\raggedright TabFact & \raggedright Accuracy & \raggedright 100 from source train & \raggedright 500 from source test \tabularnewline
\bottomrule
\end{tabular}
\end{table}
\FloatBarrier

No separate development set is used for hyperparameter tuning or checkpoint selection in the main evaluation. The source splits and the study-specific allocations are distinguished explicitly. For MMLU-Pro, the supplied allocation uses examples from the benchmark's source test split for both study training and study evaluation, with separate sampled indices assigned to the two uses. HumanEval reuses the same 100 training examples as LiveCodeBench; its 164 test examples are evaluated separately.

\paragraph{Main-table sampling.}

For MMLU-Pro, 600 indices are drawn without replacement from the source test split using Python \texttt{random.Random(42)}. The sampled indices are then sorted by source index. The first 500 selected indices are assigned to evaluation and the last 100 to training.

StrategyQA is formed by concatenating the development and training partitions in \texttt{dev + train} order and sampling 600 examples with seed 42. The sampled examples are sorted by their indices in the concatenated pool; the first 500 are assigned to evaluation and the last 100 to training.

AQuA-RAT, GSM8K, and TAT-QA each use 100 training examples selected with seed 42 and their complete stated test splits. LiveCodeBench evaluation uses all 175 examples with \texttt{version\_\allowbreak{}tag="v6"} in the cumulative \texttt{release\_\allowbreak{}v6} collection. After excluding this test subset, 100 examples are sampled from the remaining collection with seed 42 for training. HumanEval reuses these same 100 training examples and evaluates on all 164 HumanEval test examples. TabFact uses 100 source training examples and 500 source test examples, selected with seed 42. \methodname{} uses random initialization with seed 42 and no behavior-cloning initialization.

For Q1, each dataset-specific policy is trained for one pass over its 100 training examples. Each example yields two trajectories and one optimizer update, giving 200 trajectories and 100 updates. The policy sampling temperature decreases linearly from 1.2 to 1.0 over these updates, and evaluation uses the final policy weights. Q4 reuses these trained weights and the same training and evaluation subsets as Q1. All random selections for the main evaluation use sampling without replacement with seed 42. Unless an explicit release is specified above, datasets use the latest versions available when the experiments were conducted.

\paragraph{Dataset-specific training and utility probes.}

A separate construction policy is trained for each dataset. Training queries, utility probes, and final evaluation examples serve distinct roles. Four probe questions are selected without replacement from the remaining training pool using a reproducible, seed-dependent pseudorandom hash ordering, and are held fixed throughout each trajectory group. The current training query is excluded. Gold answers and code tests are used for scoring after execution and are excluded from policy observations.

\paragraph{Baseline scope.}
\label{app:baseline_protocols}

\paragraph{Baseline results and provenance.}

The main table compares \methodname{} with 17 baselines whose results are taken from \emph{OpenMAS-GCom: A Comprehensive Benchmark for Graph-Enhanced Multi-Agent Systems}. We refer readers to that work for the baseline implementations, configurations, prompts, and evaluation protocols. The methods are grouped into the following three categories.

\begin{table}[!ht]
\centering
\small
\setlength{\tabcolsep}{3pt}
\renewcommand{\arraystretch}{1.12}
\caption{Baseline categories reported in OpenMAS-GCom.}
\label{app:tab:baseline_categories}
\begin{tabular}{@{}p{0.36\linewidth}p{0.54\linewidth}@{}}
\toprule
\raggedright Category & \raggedright Methods \tabularnewline
\midrule
\raggedright Direct / single agent & \raggedright DeepSeek V4 Flash; Qwen Flash \tabularnewline
\raggedright Established MAS & \raggedright AutoGen; LLM-Debate \tabularnewline
\raggedright Adaptive / learned organization & \raggedright DyLAN; GPTSwarm; G-Designer; Sparse-Comm; GraphSearch; Graph-R1; R-GFM; BIGMAS; MasRouter; GoAgent; VeriMAP; ARG-Designer; EIB-Learner \tabularnewline
\bottomrule
\end{tabular}
\end{table}
\FloatBarrier

\methodname{} uses the initial release of \texttt{deepseek-v4-flash} for Q1. Baseline configurations are documented in OpenMAS-GCom.

\paragraph{Q4 result reuse and token accounting.}
The efficiency analysis reuses Q1 performance results on MMLU-Pro, GSM8K,
HumanEval, and TabFact and compares them with inference-token consumption.
Performance scores and token measurements are obtained from the same execution records.
It does not introduce a separate training objective or new dataset allocation.
Inference performance–cost measurements count both input and output tokens across all LLM calls. Accounting includes failed attempts, retries, repair calls, and every LLM turn before and after tool execution, as well as group-internal nodes and the outer finalizer. Internal recovery costs are therefore included alongside successful calls. Training-resource accounting is kept separate from final inference consumption.

\subsection{Component Ablations (Q2)}
\label{app:q2_protocol}

\paragraph{Component comparisons.}

On MMLU-Pro and TAT-QA, All-Atomic and All-Group restrict realizations during both training and inference. Initialization only evaluates the policy before optimization. Final-reward only retains the mixed action space and RL training while removing dense utility and structural rewards. These four variants are compared with Full using Qwen Flash on the same evaluation examples within this experiment. MMLU-Pro uses accuracy and TAT-QA uses official F1. These allocations are specific to Q2 and are not inherited from the Q1 data table.

\paragraph{Data partition and training.}
For MMLU-Pro, we read the locally cached official test split in its original
file order and shuffle the rows using Python's \texttt{random.Random(23)}.
The first 100 questions are used for evaluation and the next 100 for policy
training. These are disjoint subsets constructed within the official test
split, without subject stratification or filtering by difficulty or model
performance. For TAT-QA, we partition the locally cached official development
split by table UID, defined as the prefix before the colon in each question's
\texttt{uid}. Groups are listed in first-occurrence order and shuffled using
\texttt{random.Random(23)}. Preserving the original question order within each
group, we select 100 evaluation questions followed by 100 training questions
from subsequent, disjoint groups. Unused questions in boundary groups are
discarded. This separates tables and their associated context, rather than
entire annual reports; no stratification by question type, difficulty, or
answer category is applied.
All variants share the same partition within each dataset. Each trained
variant undergoes 100 updates with two trajectories per query;
Initialization only receives no updates.

For each dataset, Full, Final-reward only, All-Atomic, and All-Group each make one pass over the 100 training queries. Each query yields a group of two trajectories and one optimizer update, giving 200 trajectories and 100 updates per configuration. Training starts from random initialization without \textsc{stop} warmup. Evaluation uses the final parameters after update 100 in evaluation mode, with no intermediate-checkpoint selection or early stopping. Initialization only evaluates the randomly initialized parameters with zero updates. The comparison matches training-query, trajectory, and update counts; realized token consumption and API costs vary with granularity, probe executions, and ledger reuse.

Q2 uses initialization seed 23, with all configurations on a given dataset sharing the same initial policy weights. During the 100 training updates, the policy sampling temperature decreases linearly from 1.2 to 1.0. Inference uses probabilistic sampling at temperature 1.0 with dropout disabled. Apart from the specified ablations and these seed and temperature settings, the configurations share the core architecture, optimizer, regularization, graph limits, and executor settings in Table~\ref{app:tab:current_configuration}. Dense-reward configurations use four fixed probes per trajectory group and the listed shaping weights, discount, and free node/edge allowances.

\paragraph{Execution and reporting.}
Qwen Flash uses JSON-object output, temperature zero, thinking disabled,
and a 2,048-token output limit. Evaluation samples graphs at policy
temperature one with dropout disabled. Full and the fixed-granularity
variants use the potential-based reward; Final-reward only sets the three
shaping weights to zero. Reported tokens sum executor input and output usage
over the 100 evaluation questions, including evaluation retries and
excluding training calls.

\subsection{Optimization Strategy Comparison (Q3)}
\label{app:q3_protocol}

\paragraph{Data and sample selection.}
Evaluation uses 200 MMLU-Pro source-test queries and 200 TAT-QA source-dev
queries. For each dataset, the original training experiment's 100 training
examples and 100 evaluation examples are excluded; TAT-QA exclusions apply
to the corresponding complete table UIDs. Remaining IDs are sorted, shuffled
with seed 23, and the first 200 are selected. The selected memberships are
recorded in the dataset-specific \texttt{membership.json} files.

\paragraph{Comparison and training budget.}
Both routes cycle through a fixed set of 100 training questions per dataset
and start from identical random weights with seed 23. They use a one-layer,
256-dimensional edge-aware GRU with training dropout 0.1, at most three
outer units and expanded depth four, and Adam with learning rate $10^{-4}$.
Both start from the same ledger snapshot and subsequently maintain separate
caches; cached graph--question scores incur no additional API cost.
The comparison controls Qwen Flash API expenditure, not parameter-update
counts. Checkpoints are saved at initialization and CNY 2, 4, 6, 8, and 10
budget stages. Costs use call usage and official pricing, with complete-batch
accounting allowing small budget overshoots; evaluation costs are separate.

\paragraph{Search and demonstration selection.}
For each question, Search-then-SFT generates eight distinct candidates,
including the empty graph, using seeded random roles and granularities over
single-node, two-node serial/parallel, three-node converging, and random legal
DAG shapes. Candidates are generated independently of the policy network.
Scored candidates accumulate in a per-question pool. If the best score is
positive, one highest-scoring graph is subjected to legal single-node and
single-edge deletions, which are scored and added to the pool. Up to four
highest-scoring positive candidates are retained, breaking ties by graph hash.
Thus MMLU-Pro demonstrations are correct answers, whereas TAT-QA
demonstrations have the highest positive F1 in the pool, not necessarily F1 one.

\paragraph{Stage-wise SFT and RL updates.}
At each nonzero budget stage, SFT runs one epoch over questions with a
demonstration, selecting one graph per question and rotating tied
demonstrations across stages. Each graph is serialized into node-addition
actions followed by \textsc{stop}. One Adam update per question minimizes
role and granularity cross-entropy plus legal-predecessor binary
cross-entropy, averaged over action positions including \textsc{stop}.
Demonstrations may be reused across stages. SFT incurs no LLM API calls
and uses no RL KL or entropy regularizers; search resumes after each stage.
RL instead samples two current-policy trajectories per question and updates
once per group. It uses the final potential-based reward with four shared
fixed probes, discount one, shaping weights $(1,0.02,0.05)$, node/edge free
allowances of six, and zero terminal call/token penalties. Its KL and entropy
coefficients are 0.01 and 0.001.

On MMLU-Pro, Search-then-SFT performs 411 search batches and 423 SFT updates
at CNY 10.0121, while RL performs 215 updates at CNY 10.0748. On TAT-QA,
the corresponding totals are 488 search batches and 458 SFT updates at
CNY 10.0104, versus 194 RL updates at CNY 10.0497.

\paragraph{Evaluation settings.}
Execution uses Qwen Flash with \texttt{json\_object}, thinking disabled,
temperature zero, and a 2,048-token output limit. Graph construction is
probabilistic with dropout disabled; seed 23 and the query ID determine the
construction randomness. The API concurrency limit is eight. The common
initial policy is evaluated once per dataset and its results are shown for
both routes. Across all stages, evaluation comprises 4,400 actual MAS
executions and 4,800 reported rows; utility-ledger answers are not reused.

\paragraph{Scoring and accounting.}
MMLU-Pro uses accuracy; TAT-QA uses official F1, with EM recorded separately.
Valid final-answer scores and execution-failure events are recorded separately.
Evaluation-token records include inputs, outputs, post-tool turns,
summarization, and retries with returned usage. Shared initialization is
displayed under both routes but counted only once in physical execution totals.
These evaluation tokens are distinct from the training API cost on the
budget axis. Results are aggregated over the 200 evaluation examples per
dataset at each saved stage.

\subsection{Implementation and Reproducibility}
\label{app:configuration}

\paragraph{Reference implementation settings and run-specific overrides.}

Table~\ref{app:tab:current_configuration} summarizes the core configuration used in Q1 and Q2 Full. These runs share the policy architecture, optimizer and learning rate, regularization, reward settings, graph limits, probe count, and executor decoding and output settings. Q1 uses initialization seed 42 and DeepSeek V4 Flash, while Q2 uses seed 23 and Qwen Flash. Q4 reuses the Q1 policies and configuration. The Q2 ablations modify the components specified above.
Q3 uses Qwen Flash under its separate protocol. D1's expanded-depth limit is six and D3's diagnostic group size is eight.

\begin{table}[!ht]
\centering
\small
\setlength{\tabcolsep}{3pt}
\renewcommand{\arraystretch}{1.12}
\caption{Reference implementation settings.}
\label{app:tab:current_configuration}
\begin{tabular}{@{}p{0.36\linewidth}p{0.54\linewidth}@{}}
\toprule
\raggedright Component & \raggedright Setting \tabularnewline
\midrule
\raggedright Frozen semantic encoder & \raggedright \texttt{sentence-transformers/all-MiniLM-L6-v2}, 384 dimensions \tabularnewline
\raggedright Graph encoder & \raggedright One shared edge-aware GRUCell, hidden size 256 \tabularnewline
\raggedright GRU input dropout & \raggedright 0.1 during training; disabled at inference \tabularnewline
\raggedright Maximum outer units / expanded depth & \raggedright 3 / 4; outer summarizer excluded from depth \tabularnewline
\raggedright Q1/Q2 training policy temperature & \raggedright Linear annealing from 1.2 to 1.0 over 100 updates \tabularnewline
\raggedright Inference policy temperature & \raggedright 1.0; probabilistic sampling \tabularnewline
\raggedright Initialization & \raggedright Random; seed 42 for Q1, 23 for Q2; no behavior-cloning initialization or \textsc{stop} warmup \tabularnewline
\raggedright RL trajectories per query group & \raggedright 2 \tabularnewline
\raggedright Fixed utility probes per group & \raggedright 4 other training queries \tabularnewline
\raggedright Optimizer / learning rate & \raggedright Adam with default parameters / fixed $10^{-4}$ \tabularnewline
\raggedright KL / entropy coefficients & \raggedright 0.01 / 0.001 \tabularnewline
\raggedright Advantage epsilon & \raggedright $10^{-8}$, scaled consistently during numerical normalization \tabularnewline
\raggedright Reward weights $(\alpha,\beta,\gamma_R)$ & \raggedright $(1,0.02,0.05)$ \tabularnewline
\raggedright Shaping and return discount & \raggedright 1 \tabularnewline
\raggedright Utility smoothing $(\kappa,u_0)$ & \raggedright $(1,0.5)$ \tabularnewline
\raggedright Expanded node / edge weights & \raggedright 1 / 1 \tabularnewline
\raggedright Free expanded node / edge allowances & \raggedright 6 / 6 \tabularnewline
\raggedright Terminal call / token penalty weights & \raggedright 0 / 0 \tabularnewline
\raggedright Executor temperature / thinking & \raggedright 0 / disabled \tabularnewline
\raggedright Executor output mode / output limit & \raggedright \texttt{json\_\allowbreak{}object} / 2,048 tokens \tabularnewline
\bottomrule
\end{tabular}
\end{table}
\FloatBarrier

The results of our Q1, Q2, and Q4 evaluations are obtained from a single run per evaluated configuration, with one complete system execution per evaluation question. Reported benchmark scores aggregate question-level scores within that run.

Adam uses default parameters apart from the learning rate, with no gradient clipping or learning-rate schedule. Policy sampling and executor decoding use separate temperatures. The unit horizon and depth bound constrain organization structure; the free node and edge allowances define the potential. The reward configuration's call and token normalizers do not impose execution hard limits.

The reference policy is frozen before the first RL update and remains in evaluation mode. The current policy's regularizers reuse the sampling dropout realization, including RNG replay for \textsc{stop}, as detailed in Appendix~\ref{app:implementation_training_details}. Reproducibility therefore depends on policy sampling and dropout randomness as well as the dataset and probe selection procedures.

\paragraph{Run count and aggregation.}
Given the API cost, Q1, Q2, and Q3 use one run per dataset and experimental
condition. Reported scores aggregate evaluation examples within that run,
rather than averaging across independent training seeds. Q3's budget stages
are checkpoints along the same run for each route, not independent repeats.

Q1 uses seed 42 and Q2 uses seed 23 for policy initialization, policy sampling, and training dropout. D3 uses eight trajectories per query under its separate diagnostic protocol.

\paragraph{Software, hardware, and resource measurement.}

\subsection{Role Libraries, Prompts, and Execution Interfaces}
\label{app:role_protocols}

\paragraph{Role library and realizations.}

The documented framework registers 14 top-level roles, each with an atomic and a group realization. A dataset profile selects the available roles and tools; the construction policy selects the units that actually execute. An atomic realization uses one executor. Each group contains three workers feeding one internal aggregator, with no direct worker-to-worker edges. Table~\ref{app:tab:role_library} lists the registered roles and their internal assignments. These library definitions remain fixed during policy training and inference.

\begin{table}[!ht]
\centering
\small
\setlength{\tabcolsep}{3pt}
\renewcommand{\arraystretch}{1.12}
\caption{Registered roles and group realizations.}
\label{app:tab:role_library}
\begin{tabular}{@{}p{0.2\linewidth}p{0.22\linewidth}p{0.28\linewidth}p{0.2\linewidth}@{}}
\toprule
\raggedright Top-level role & \raggedright Atomic responsibility & \raggedright Three group workers & \raggedright Internal aggregator \tabularnewline
\midrule
\raggedright \texttt{task\_\allowbreak{}decomposer} & \raggedright Identify constraints, subproblems, dependencies, and output requirements & \raggedright \texttt{constraint\_\allowbreak{}extractor}; \texttt{subproblem\_\allowbreak{}planner}; \texttt{dependency\_\allowbreak{}checker} & \raggedright \texttt{decomposition\_\allowbreak{}aggregator} \tabularnewline
\raggedright \texttt{evidence\_\allowbreak{}analyst} & \raggedright Extract, rank, and reconcile public evidence & \raggedright \texttt{evidence\_\allowbreak{}finder}; \texttt{relevance\_\allowbreak{}ranker}; \texttt{contradiction\_\allowbreak{}finder} & \raggedright \texttt{evidence\_\allowbreak{}aggregator} \tabularnewline
\raggedright \texttt{independent\_\allowbreak{}solver} & \raggedright Solve independently and assess predecessor conclusions & \raggedright \texttt{solver\_\allowbreak{}a}; \texttt{solver\_\allowbreak{}b}; \texttt{assumption\_\allowbreak{}checker} & \raggedright \texttt{solution\_\allowbreak{}consensus} \tabularnewline
\raggedright \texttt{adversarial\_\allowbreak{}verifier} & \raggedright Check counterexamples, consistency, omissions, and answer format & \raggedright \texttt{counterexample\_\allowbreak{}searcher}; \texttt{consistency\_\allowbreak{}checker}; \texttt{format\_\allowbreak{}checker} & \raggedright \texttt{verification\_\allowbreak{}aggregator} \tabularnewline
\raggedright \texttt{knowledge\_\allowbreak{}reasoner} & \raggedright Apply model knowledge and identify uncertainty & \raggedright \texttt{fact\_\allowbreak{}recaller}; \texttt{contextual\_\allowbreak{}reasoner}; \texttt{uncertainty\_\allowbreak{}calibrator} & \raggedright \texttt{knowledge\_\allowbreak{}aggregator} \tabularnewline
\raggedright \texttt{option\_\allowbreak{}eliminator} & \raggedright Assess support and objections for candidate options & \raggedright \texttt{option\_\allowbreak{}support\_\allowbreak{}analyst}; \texttt{option\_\allowbreak{}refute\_\allowbreak{}analyst}; \texttt{distractor\_\allowbreak{}detector} & \raggedright \texttt{option\_\allowbreak{}aggregator} \tabularnewline
\raggedright \texttt{logic\_\allowbreak{}reasoner} & \raggedright Formalize premises and perform logical, causal, and Boolean reasoning & \raggedright \texttt{premise\_\allowbreak{}formalizer}; \texttt{forward\_\allowbreak{}reasoner}; \texttt{countermodel\_\allowbreak{}searcher} & \raggedright \texttt{logic\_\allowbreak{}aggregator} \tabularnewline
\bottomrule
\end{tabular}
\end{table}
\FloatBarrier
\begin{table}[!ht]
\centering
\small
\setlength{\tabcolsep}{3pt}
\renewcommand{\arraystretch}{1.12}
\caption{Registered roles and group realizations. (continued).}
\begin{tabular}{@{}p{0.2\linewidth}p{0.22\linewidth}p{0.28\linewidth}p{0.2\linewidth}@{}}
\toprule
\raggedright Top-level role & \raggedright Atomic responsibility & \raggedright Three group workers & \raggedright Internal aggregator \tabularnewline
\midrule
\raggedright \texttt{quantitative\_\allowbreak{}modeler} & \raggedright Define variables, equations, units, and a solution plan & \raggedright \texttt{variable\_\allowbreak{}mapper}; \texttt{equation\_\allowbreak{}builder}; \texttt{unit\_\allowbreak{}constraint\_\allowbreak{}checker} & \raggedright \texttt{model\_\allowbreak{}aggregator} \tabularnewline
\raggedright \texttt{calculation\_\allowbreak{}auditor} & \raggedright Recompute numerical results and identify discrepancies & \raggedright \texttt{primary\_\allowbreak{}calculator}; \texttt{independent\_\allowbreak{}recalculator}; \texttt{magnitude\_\allowbreak{}unit\_\allowbreak{}checker} & \raggedright \texttt{calculation\_\allowbreak{}aggregator} \tabularnewline
\raggedright \texttt{algorithm\_\allowbreak{}designer} & \raggedright Develop algorithms and assess complexity and boundary cases & \raggedright \texttt{algorithm\_\allowbreak{}candidate\_\allowbreak{}a}; \texttt{algorithm\_\allowbreak{}candidate\_\allowbreak{}b}; \texttt{complexity\_\allowbreak{}edgecase\_\allowbreak{}analyst} & \raggedright \texttt{algorithm\_\allowbreak{}aggregator} \tabularnewline
\raggedright \texttt{code\_\allowbreak{}implementer} & \raggedright Implement the required function or input/output interface & \raggedright \texttt{implementation\_\allowbreak{}drafter}; \texttt{interface\_\allowbreak{}specialist}; \texttt{public\_\allowbreak{}sample\_\allowbreak{}runner} & \raggedright \texttt{code\_\allowbreak{}aggregator} \tabularnewline
\raggedright \texttt{code\_\allowbreak{}reviewer} & \raggedright Review correctness, complexity, and boundary behavior & \raggedright \texttt{static\_\allowbreak{}analyzer}; \texttt{edge\_\allowbreak{}case\_\allowbreak{}generator}; \texttt{complexity\_\allowbreak{}reviewer} & \raggedright \texttt{review\_\allowbreak{}aggregator} \tabularnewline
\raggedright \texttt{table\_\allowbreak{}interpreter} & \raggedright Locate rows, columns, entities, time references, and scales & \raggedright \texttt{row\_\allowbreak{}locator}; \texttt{column\_\allowbreak{}relation\_\allowbreak{}analyst}; \texttt{temporal\_\allowbreak{}scale\_\allowbreak{}checker} & \raggedright \texttt{table\_\allowbreak{}aggregator} \tabularnewline
\raggedright \texttt{entailment\_\allowbreak{}judge} & \raggedright Assess a claim using supporting and refuting evidence & \raggedright \texttt{support\_\allowbreak{}case\_\allowbreak{}builder}; \texttt{refutation\_\allowbreak{}case\_\allowbreak{}builder}; \texttt{quantifier\_\allowbreak{}scope\_\allowbreak{}checker} & \raggedright \texttt{entailment\_\allowbreak{}aggregator} \tabularnewline
\bottomrule
\end{tabular}
\end{table}
\FloatBarrier

Role responsibilities are specified through prompts. Intermediate roles may supply a complete candidate answer or \texttt{null} when unable to answer. Aggregators combine their workers' contributions, retaining relevant disagreements, uncertainty, and failed checks. Roles named “finder” or “searcher” examine public task information or model knowledge; the documented tool registry contains no web-search tool.

\paragraph{Tool access.}

The executor supports four tools. \texttt{calculator} evaluates restricted expressions; \texttt{python\_\allowbreak{}exec} runs a standalone Python program in a sandbox and returns stdout, stderr, and exit status; \texttt{public\_\allowbreak{}code} performs public code checks and sample tests; and \texttt{table\_\allowbreak{}reader} accesses the task's public table data. Public code tests use either standard-input/output comparisons or function-call comparisons with supplied expected outputs. Hidden benchmark tests remain part of final scoring.

\begin{table}[!ht]
\centering
\small
\setlength{\tabcolsep}{3pt}
\renewcommand{\arraystretch}{1.12}
\caption{Executor-level tool permissions.}
\label{app:tab:tool_permissions}
\begin{tabular}{@{}p{0.36\linewidth}p{0.54\linewidth}@{}}
\toprule
\raggedright Atomic role or group worker & \raggedright Permitted tools \tabularnewline
\midrule
\raggedright \texttt{calculation\_\allowbreak{}auditor}, \texttt{primary\_\allowbreak{}calculator}, \texttt{independent\_\allowbreak{}recalculator}, \texttt{magnitude\_\allowbreak{}unit\_\allowbreak{}checker} & \raggedright \texttt{calculator}, \texttt{python\_\allowbreak{}exec} \tabularnewline
\raggedright \texttt{code\_\allowbreak{}implementer}, \texttt{public\_\allowbreak{}sample\_\allowbreak{}runner} & \raggedright \texttt{public\_\allowbreak{}code}, \texttt{python\_\allowbreak{}exec} \tabularnewline
\raggedright \texttt{code\_\allowbreak{}reviewer}, \texttt{static\_\allowbreak{}analyzer}, \texttt{edge\_\allowbreak{}case\_\allowbreak{}generator} & \raggedright \texttt{public\_\allowbreak{}code} \tabularnewline
\raggedright \texttt{table\_\allowbreak{}interpreter}, \texttt{row\_\allowbreak{}locator}, \texttt{column\_\allowbreak{}relation\_\allowbreak{}analyst}, \texttt{temporal\_\allowbreak{}scale\_\allowbreak{}checker} & \raggedright \texttt{table\_\allowbreak{}reader} \tabularnewline
\bottomrule
\end{tabular}
\end{table}
\FloatBarrier

Other registered execution roles have no tool permission. Actual availability is the intersection of executor-level permissions and tools enabled by the dataset profile, subject to runtime availability. Each group worker has its own permissions; atomic-role permissions are not automatically inherited by the group members or aggregator.

\paragraph{Content and tool-request protocol.}

The documented content protocol is \texttt{analysis-answer-v2}. Its core object requires \texttt{analysis} and \texttt{answer}, with additional fields permitted:

\begingroup
\small
\begin{verbatim}
{
  "analysis": "Brief reasoning, evidence, checks, and uncertainty",
  "answer": "Complete candidate answer"
}
\end{verbatim}
\endgroup

\texttt{analysis} must be a nonempty string. \texttt{answer} follows the dataset-specific contract in Table~\ref{app:tab:answer_contracts} and may be \texttt{null} when a node cannot answer. The model returns JSON without Markdown fences or a \texttt{FINAL:} prefix. Ordinary atomic nodes, group workers, and internal aggregators wrap their content in a node-response envelope:

\begingroup
\small
\begin{verbatim}
{
  "kind": "content",
  "content": {
    "analysis": "Checked the public evidence",
    "answer": "42"
  }
}
\end{verbatim}
\endgroup

A permitted tool is requested with the alternative envelope:

\begingroup
\small
\begin{verbatim}
{
  "kind": "tool_request",
  "tool_request": {
    "tool": "python_exec",
    "payload": {
      "source": "print((120 - 100) / 100 * 100)"
    }
  }
}
\end{verbatim}
\endgroup

Each node execution permits at most one tool request. After the executor supplies the actual tool result, the next model response must submit content. A normal execution therefore uses one model response without a tool or two with a tool; intermediate-node repair, truncation recovery, and transport retries may add API calls under the rules in Appendix~\ref{app:datasets_scoring}. The runtime supplies tool observations, status, timing, and token-accounting fields. These fields are not generated by the model.

\paragraph{Routing and finalization.}

Parsed outputs are wrapped in runtime packets carrying provenance, task association, and execution status. Downstream requests contain the public task, role responsibilities, answer contract, and projected messages from graph-specified predecessors. Message projection and length budgets determine the content delivered to each node. Gold answers and hidden tests are excluded from solving inputs.

All three group workers receive every external direct predecessor's result. The internal aggregator receives the three workers as its direct predecessors and emits the group's external result. The outer finalizer receives the public question, output contract, and packets from all sinks of the expanded graph. The finalizer and the documented direct single-LLM path emit the core \texttt{analysis}/\texttt{answer} object without the ordinary node's \texttt{kind} envelope. Appendix~\ref{app:implementation_training_details} provides the construction-to-execution flow.

\paragraph{Dataset answer contracts.}

\begin{table}[!ht]
\centering
\small
\setlength{\tabcolsep}{3pt}
\renewcommand{\arraystretch}{1.12}
\caption{Required model answer formats.}
\label{app:tab:answer_contracts}
\begin{tabular}{@{}p{0.36\linewidth}p{0.54\linewidth}@{}}
\toprule
\raggedright Dataset & \raggedright \texttt{answer} format \tabularnewline
\midrule
\raggedright MMLU-Pro & \raggedright Option-label string, e.g., \texttt{"A"} \tabularnewline
\raggedright AQuA-RAT & \raggedright Option-label string, e.g., \texttt{"B"} \tabularnewline
\raggedright StrategyQA & \raggedright String \texttt{"true"} or \texttt{"false"} \tabularnewline
\raggedright TabFact & \raggedright String \texttt{"true"} or \texttt{"false"} \tabularnewline
\raggedright GSM8K & \raggedright Numeric string without units, currency symbols, or commas; integers omit \texttt{.0} \tabularnewline
\raggedright HumanEval & \raggedright Complete function source as a string without code fences \tabularnewline
\raggedright LiveCodeBench-v6 & \raggedright Complete code string matching the problem interface, without code fences \tabularnewline
\raggedright TAT-QA & \raggedright Object with \texttt{values} and \texttt{scale} fields \tabularnewline
\bottomrule
\end{tabular}
\end{table}
\FloatBarrier

StrategyQA and TabFact use string values rather than JSON Booleans. For TAT-QA, each answer span is a separate string in \texttt{values}; arithmetic and counting answers contain a single numeric value as required by the prompt. Allowed scales are the empty string, \texttt{thousand}, \texttt{million}, \texttt{billion}, and \texttt{percent}. Numeric answers do not repeat unit suffixes. For example:

\begingroup
\small
\begin{verbatim}
{
  "analysis": "Compute the growth rate from the relevant years",
  "answer": {
    "values": [
      "15.64"
    ],
    "scale": "percent"
  }
}
\end{verbatim}
\endgroup

The runtime deterministically converts the validated final \texttt{answer} into the representation required by the scorer, including its internal \texttt{FINAL:} wrapper. The model itself does not emit this wrapper, and conversion requires no model call. Format validation and task scoring are separate: the scorer evaluates the answer's numerical, textual, and scale correctness after conversion.

\paragraph{Prompt and profile artifacts.}

The tables and protocol examples above specify the registered library and communication contracts. Exact role prompts and dataset-specific profiles are separate reproducibility artifacts.

\section{Extended Empirical Investigation}
\label{app:extended_empirical_statistics}

This section supplements the three diagnostics in the main-text Empirical Investigation: local granularity preferences, verified configuration coverage, and construction-level credit.

\subsection{Controlled Granularity Assignments}
\label{app:diagnostic_d1}

\paragraph{Protocol.}

We fix a serial decomposer–solver–verifier skeleton and enumerate its eight atomic/group assignments: AAA, AAG, AGA, AGG, GAA, GAG, GGA, and GGG. The letters follow the role order; A denotes an atomic realization and G a group realization. All assignments use the same 200 MMLU-Pro test queries and 200 TAT-QA development queries, with one run per query and assignment. The role library, execution model, prompts, and decoding settings remain fixed across assignments. This gives 1,600 query–configuration executions per dataset.

For each dataset, the adapter produces a stable \texttt{example\_\allowbreak{}id}. Examples are sorted by the ascending SHA-256 digest below, and the first 200 are selected:

\begingroup
\small
\begin{verbatim}
key = hashlib.sha256(f"d1-v1\0{42}\0{example_id}".encode()).digest()
\end{verbatim}
\endgroup

This seed-42 selection is unstratified. TAT-QA is sampled at the question level, rather than by table or document. All eight assignments share the selected examples. D1 explicitly sets the maximum expanded depth to six, rather than the depth-four limit of the current construction policy, so all eight serial assignments, including GGG, are admissible. The diagnostic reports accuracy and total input/output tokens on these diagnostic-specific sets.

\paragraph{Historical TAT-QA scoring.}

D1 uses a custom binary scorer that extracts an answer string from a unique terminal \texttt{FINAL:} marker. This historical protocol precedes the official F1 scorer and the current \texttt{values}/\texttt{scale} object contract. Its principal rules are:

\begin{table}[!ht]
\centering
\small
\setlength{\tabcolsep}{3pt}
\renewcommand{\arraystretch}{1.12}
\caption{D1 TAT-QA binary scoring rules.}
\label{app:tab:d1_scoring}
\begin{tabular}{@{}p{0.22\linewidth}p{0.7\linewidth}@{}}
\toprule
\raggedright Answer type & \raggedright Correctness rule \tabularnewline
\midrule
\raggedright Arithmetic & \raggedright Parse prediction and reference as Decimal values, round both to two decimal places using ties-to-even, and compare for equality. \tabularnewline
\raggedright Count & \raggedright Compare Decimal values exactly, without two-decimal rounding. \tabularnewline
\raggedright Numeric syntax and scale & \raggedright Accept thousands separators, leading currency symbols, and parentheses for negatives. A nonempty reference scale requires a matching \texttt{percent}, \texttt{thousand}, \texttt{million}, or \texttt{billion} suffix; \texttt{\%} is also accepted for percent. No cross-scale conversion is applied. \tabularnewline
\raggedright Single text span & \raggedright Apply Unicode NFKC and case normalization, collapse whitespace, strip designated boundary punctuation, and require exact equality. \tabularnewline
\raggedright Multiple spans & \raggedright Split on semicolons, normalize each span, and compare multisets using \texttt{Counter}, ignoring order while retaining multiplicity. \tabularnewline
\raggedright Numeric span & \raggedright Apply numeric normalization and two-decimal comparison; compare scale as well for scaled numeric spans. \tabularnewline
\bottomrule
\end{tabular}
\end{table}
\FloatBarrier

The recorded correctness also requires execution without a failure event:

\begingroup
\small
\begin{verbatim}
correct = score.correct and not result.failures
\end{verbatim}
\endgroup

Consequently, a correct answer accompanied by a recorded MAS execution failure receives zero credit in D1. This diagnostic-specific rule is distinct from the final-answer scoring protocol in Appendix~\ref{app:experimental_details}, which records internal failures separately.

\paragraph{Role-conditioned comparisons.}

To isolate a local realization change, we compare a mixed assignment with AAA while keeping the other two roles atomic. The final-answer accuracy changes reported in the main text are:

\begin{table}[!ht]
\centering
\small
\setlength{\tabcolsep}{3pt}
\renewcommand{\arraystretch}{1.12}
\caption{D1 role-level granularity interventions.}
\label{app:tab:d1_interventions}
\begin{tabular}{@{}p{0.4\linewidth}p{0.25\linewidth}p{0.25\linewidth}@{}}
\toprule
\raggedright Intervention relative to AAA & \raggedright MMLU-Pro change (percentage points) & \raggedright TAT-QA change (percentage points) \tabularnewline
\midrule
\raggedright Group only the decomposer: GAA & \raggedright +4.5 & \raggedright 0.0 \tabularnewline
\raggedright Group only the verifier: AAG & \raggedright $-$1.0 & \raggedright +2.5 \tabularnewline
\bottomrule
\end{tabular}
\end{table}
\FloatBarrier

These differences concern the diagnostic correctness measure after the full fixed skeleton executes. MMLU-Pro favors GAA, reaching 81.5\% accuracy versus 77.0\% for both fixed endpoints, whereas TAT-QA favors AAG, reaching 70.0\% versus 67.5\%. Relative to GGG, these mixed assignments reduce token consumption by 58.72\% and 48.15\%, respectively. On each dataset, AAA and the preferred mixed assignment form the observed accuracy–token frontier: the mixed assignment increases accuracy at additional cost over AAA, while using fewer tokens than GGG. The controlled assignments therefore identify task-dependent choices about where to allocate group execution.

\subsection{Verified Configuration Coverage}
\label{app:diagnostic_d2}

Fix the role sequence and outer dependency skeleton, with $m$ positions that each admit both atomic and group realizations. The diagnostic counts their binary realization assignments:

\begin{equation}
\mathcal Z_{\mathrm{mix}}=\{A,G\}^{m},\qquad
|\mathcal Z_{\mathrm{mix}}|=2^m.
\label{app:eq:diagnostic_1}
\end{equation}

Let $V_B$ be the set of distinct assignments verified under a budget of $B$ complete graph executions. Since each verification consumes at least one such execution,

\begin{equation}
|V_B|\le \min(2^m,B),\qquad
\operatorname{Coverage}(B)=\frac{|V_B|}{2^m}
\le \min\left(1,\frac{B}{2^m}\right).
\label{app:eq:diagnostic_2}
\end{equation}

Repeated executions of an already verified assignment consume budget without adding a distinct assignment to $V_B$. The denominator counts the binary assignments for the fixed skeleton; additional role orders and edge choices expand the construction space. This counting argument relates verified configuration coverage to graph-execution budget.

\subsection{Construction Credit on Shared Trajectories}
\label{app:diagnostic_d3}

\paragraph{Data and diagnostic updates.}

The Qwen Flash diagnostic uses 30 distinct queries per dataset from
MMLU-Pro validation, GSM8K train, and HumanEval test. Each dataset has an
independently initialized policy trained with the model and reward defined in
Section~\ref{sec:methodology}, rather than restored from an earlier diagnostic
checkpoint. Each query is used for one update with eight trajectories, giving
30 updates and 240 trajectories per dataset, or 720 trajectories overall.
HumanEval test examples serve as diagnostic training data in this protocol.
Each query group shares four fixed probes drawn from other training queries.
The eight-trajectory group size is diagnostic-specific.

\paragraph{Diagnostic return definitions.}

Terminal-only and dense returns are computed on the same recorded trajectories.
Dense rewards follow Section~\ref{sec:method_optimization} and
Appendix~\ref{app:reward_specification}, including full potential settlement
at both sampled and horizon-forced \textsc{stop}. With discount $\eta=1$,
the two return-to-go definitions are

\begin{equation}
G^{\mathrm{terminal}}_{i,t}=S_i,\qquad
G^{\mathrm{dense}}_{i,t}=S_i-\Phi_{i,t}.
\label{app:eq:diagnostic_3}
\end{equation}

Here $S_i$ is the terminal task score and $\Phi_{i,t}$ is the potential before
action $t$. Both initial returns equal $S_i$ because $\Phi_{i,0}=0$.
The implementation checks that each complete shaped return equals its
terminal score and that both initial returns agree.

\paragraph{Non-zero action advantages.}

Within each eight-trajectory query group, returns are normalized at each action position over the trajectories that contain that position:

\begin{equation}
A_{i,t}=\frac{G_{i,t}-\mu_t}{\sigma_t+10^{-8}},
\qquad \text{non-zero iff } |A_{i,t}|>10^{-8}.
\label{app:eq:diagnostic_4}
\end{equation}

The mean and standard deviation are computed separately for each return
definition. Zero-standard-deviation positions receive zero advantage.
Normalization includes all trajectories with an actual action at that position,
including \textsc{stop}; shorter trajectories contribute no padding. A
\textsc{stop} and an addition may therefore be compared at the same index.
After normalization, the primary statistic selects only intermediate additions,
$\mathcal I=\{(i,t):t>0,\ a_{i,t}\text{ is ADD}\}$, and reports
$\rho=\sum_{(i,t)\in\mathcal I}\mathbf 1[|A_{i,t}|>10^{-8}]/|\mathcal I|$.
Counts are also reported by construction position. Initial decisions,
subsequent sampled \textsc{stop}, and horizon-forced \textsc{stop} are reported
separately from this primary statistic. Empty reporting sets are marked N/A.

\paragraph{Equal-terminal-reward pairs.}

Within each update, we enumerate unordered pairs among that query's eight
trajectories. We retain equal-terminal-score pairs satisfying

\begin{equation}
|S_i-S_j|<10^{-8}.
\label{app:eq:diagnostic_5}
\end{equation}

For each retained pair, an eligible comparison position has $t>0$ and an
actual ADD action in both trajectories. A trajectory-pair--position is counted
as distinguished when

\begin{equation}
|G^{\mathrm{dense}}_{i,t}-G^{\mathrm{dense}}_{j,t}|>10^{-8}.
\label{app:eq:diagnostic_6}
\end{equation}

The denominator is the number of eligible trajectory-pair--positions, each
counted once within its query group. One pair can contribute multiple
positions. Returns are computed from the complete recorded reward sequence,
including terminal settlement; shorter trajectories are not padded.
The statistic is N/A when no eligible comparisons exist.

\paragraph{Pooled statistics.}

Table~\ref{app:tab:d3_credit_counts} reports pooled counts across updates
for each dataset. Percentages are ratios of pooled numerators and denominators,
rather than averages of per-query percentages.

\begin{table}[!ht]
\centering
\small
\setlength{\tabcolsep}{3pt}
\renewcommand{\arraystretch}{1.12}
\caption{D3 statistics under the final potential-based reward, pooled over
30 queries and 240 trajectories per dataset.}
\label{app:tab:d3_credit_counts}
\begin{tabular}{@{}p{0.16\linewidth}p{0.25\linewidth}p{0.25\linewidth}p{0.24\linewidth}@{}}
\toprule
\raggedright Dataset & \raggedright Terminal-only non-zero advantages / intermediate ADD actions & \raggedright Dense non-zero advantages / intermediate ADD actions & \raggedright Distinguished / eligible equal-score pair--positions \tabularnewline
\midrule
\raggedright MMLU-Pro & \raggedright 105/320 (32.81\%) & \raggedright 266/320 (83.13\%) & \raggedright 343/642 (53.43\%) \tabularnewline
\raggedright GSM8K & \raggedright 5/141 (3.55\%) & \raggedright 47/141 (33.33\%) & \raggedright 48/204 (23.53\%) \tabularnewline
\raggedright HumanEval & \raggedright 39/236 (16.53\%) & \raggedright 181/236 (76.69\%) & \raggedright 123/376 (32.71\%) \tabularnewline
\bottomrule
\end{tabular}
\end{table}
\FloatBarrier

Potential-based shaping supplies non-zero relative advantages
to a larger fraction of intermediate ADD actions and distinguishes some
equal-terminal-score trajectories at intermediate positions while preserving
complete task returns as established in Proposition~1.
The two action-statistic columns compare return definitions on the same
trajectory sample, rather than independent training runs.

\end{document}